%% file: main.tex
\documentclass[10pt,twocolumn,letterpaper]{article}

\usepackage[pagenumbers]{cvpr} % To force page numbers, e.g. for an arXiv version


\definecolor{cvprblue}{rgb}{0.21,0.49,0.74}
\usepackage[pagebackref,breaklinks,colorlinks,allcolors=cvprblue]{hyperref}

\def\paperID{15089} % *** Enter the Paper ID here
\def\confName{CVPR}
\def\confYear{2025}

\title{EmotiveTalk: Expressive Talking Head Generation through Audio Information Decoupling and Emotional Video Diffusion}

\author{Haotian Wang$^1$, Yuzhe Weng$^1$, Yueyan Li$^2$, Zilu Guo$^1$, JunDu$^{1}$, Shutong Niu$^1$, Jiefeng Ma$^1$, \\
Shan He$^3$, Xiaoyan Wu$^3$, Qiming Hu$^3$, BingYin$^3$, CongLiu$^3$, Qingfeng Liu$^3$ \\
$^1$ University of Science and Technology of China $^2$ Imperial College London $^3$ iFLYTEK CO.LTD.\\
{\tt\small \{az1522702192, yzweng, guozl, niust, jfma\}@mail.ustc.edu.cn, 
jundu@ustc.edu.cn, yl222@ic.ac.uk} \\ 
{\tt\small \{shanhe2, xywu10, qmhu3, bingyin, congliu2\}@iflytek.com, qfliu.research@gmail.com}
}

\usepackage{booktabs}
\usepackage{multirow}
\usepackage{float}
\usepackage{stfloats}
\usepackage[normalem]{ulem}
\usepackage{diagbox}
\useunder{\uline}{\ul}{}
\usepackage{colortbl}

\begin{document}
\twocolumn[{%
\renewcommand\twocolumn[1][]{#1}%
\maketitle
\begin{center}
    \centering
    \captionsetup{type=figure}
    \includegraphics[width=1.0\linewidth]{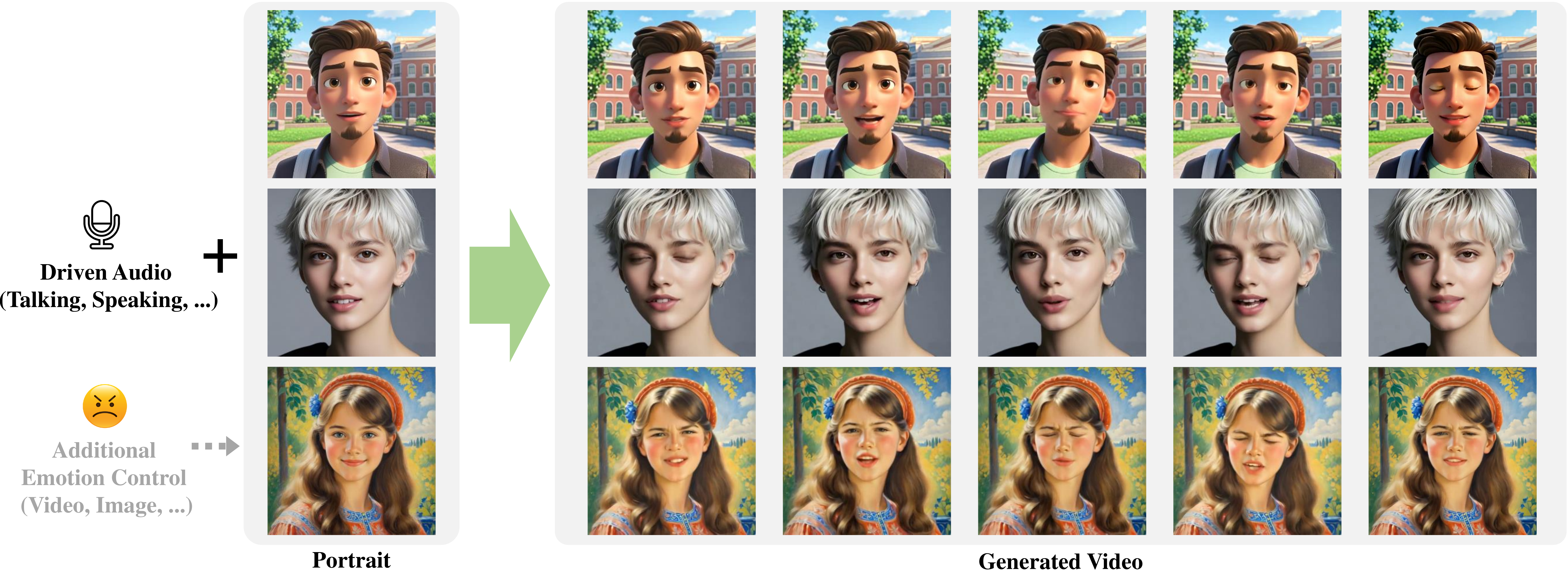}
    \caption{We propose \textbf{EmotiveTalk}, an expressive talking head generation framework. Taking a single portrait and the driven audio as input, our method can generate expressive portrait video sync with audio and custom the speaking style with additional emotion control.}
\end{center}%
}]

\input{sec/0_abstract}    
\input{sec/1_intro}
\input{sec/2_formatting}
\input{sec/3_finalcopy}
{
    \small
    \bibliographystyle{ieeenat_fullname}
    \bibliography{main}
}

\input{sec/X_suppl}
% {
%     \small
%     \bibliographystyle{ieeenat_fullname}
%     \bibliography{supplement}
% }
\end{document}

%% file: sec/0_abstract.tex
\begin{abstract}
Diffusion models have revolutionized the field of talking head generation, yet still face challenges in expressiveness, controllability, and stability in long-time generation. In this research, we propose an EmotiveTalk framework to address these issues. Firstly, to realize better control over the generation of lip movement and facial expression, a Vision-guided Audio Information Decoupling (V-AID) approach is designed to generate audio-based decoupled representations aligned with lip movements and expression. Specifically, to achieve alignment between audio and facial expression representation spaces, we present a Diffusion-based Co-speech Temporal Expansion (Di-CTE) module within V-AID to generate expression-related representations under multi-source emotion condition constraints. Then we propose a well-designed Emotional Talking Head Diffusion (ETHD) backbone to efficiently generate highly expressive talking head videos, which contains an Expression Decoupling Injection (EDI) module to automatically decouple the expressions from reference portraits while integrating the target expression information, achieving more expressive generation performance. Experimental results show that EmotiveTalk can generate expressive talking head videos, ensuring the promised controllability of emotions and stability during long-time generation, yielding state-of-the-art performance compared to existing methods. The main page of our project can be found in this url \href{https://emotivetalk.github.io/}{EmotiveTalk}.
\end{abstract}

%% file: sec/1_intro.tex
\section{Introduction}
\label{sec:introduction}

Talking head animation, also known as portrait image animation~\cite{thdsurvey}, demonstrates significant value across multiple domains, including film and television production, online education as well as human-machine interaction. The generation of realistic talking head videos involves two aspects of requirements. On the one hand, for the verbal aspect, it is essential to ensure the synchronization between speech and lip motions in the generated video~\cite{wav2lip}. On the other hand, for the non-verbal aspect, the generated video must convey non-verbal information, including vivid facial expressions~\cite{nonverbal}.

Despite the success of diffusion models~\cite{denoising,ddpm,ddim} in image and video generation tasks, their application in talking head generation~\cite{difftalk,diffusedheads,dreamtalk,emo,hallo} still faces several challenges. For example, current methodologies~\cite {dreamtalk,emo,hallo} exhibit shortcomings in control of the generated emotional facial expressions, although they have made notable advancements in achieving synchronization between speech and lip movements. These audio-driven methods mainly directly synthesize expressions under weak audio conditions~\cite{emo, hallo}. However, the coupling of multiple information embedded in audio limits the effective learning of the mapping between speech and expressions and the controllability of generated emotion. Moreover, current diffusion-based methods often struggle to generate high-resolution video due to their large scale of parameters and the associated training costs~\cite{emo,hallo}. They also face challenges in stability during long-time generation due to their auto-regressive inference strategies~\cite{dreamtalk,emo,hallo}, which can lead to error accumulation across multiple inference clips.

To address these challenges, in this paper, we introduce EmotiveTalk, a highly expressive talking head generation framework with emotion control based on video diffusion. We propose a Vision-guided Audio Information Decouple (V-AID) approach to facilitate the decoupling of lip and expression related information contained in audio signals and also the alignment of audio representations with video representations under the guidance of vision facial motion information. Specifically, to achieve better alignment between speech and expression representation spaces, we present a Diffusion-based Co-speech Temporal Expansion (Di-CTE) module, which generates temporal expression-related representations from audio under utterance emotional conditions from multiple optional driven sources. Then, to effectively drive the decoupled representations, we propose an efficient video diffusion framework for expressive talking head generation which demonstrates effectiveness and enhanced stability in talking head video generation performance. The backbone incorporates an Expression Decoupling Injector (EDI) module in our backbone to achieve the automatic decoupling of expression information from the reference portrait while facilitating the injection of expression-driven information. In summary, our contributions are as follows:
(1) We propose a Vision-Guided Audio Information Decouple (V-AID) approach that generates efficient decoupled lip-related and expression-related representations from audio for talking head generation. (2) We propose an Emotional Talking Head Diffusion (ETHD) framework that capable of efficiently generating dynamic-length videos, which achieves highly expressive talking head video generation while ensuring stability over extended durations. (3) We further enhance emotion controllability by integrating conditions from emotion-driven sources and realize the customization of generated emotions by multi-source emotion control. 

%% file: sec/2_formatting.tex
\section{Related Work}
\label{sec:relatedwork}
%-------------------------------------------------------------------------
\subsection{Audio-driven Talking Head Video Generation}
The initial focus of the audio-driven talking head video generation task was on achieving consistency between lip movements and the driving speech signal~\cite{wav2lip,difftalk}. SadTalker~\cite{sadtalker} and Audio2Head ~\cite{audio2head} integrate 3D information and control mechanisms to enhance the naturalism of head movements. Diffused Heads~\cite{diffusedheads} and DreamTalk~\cite{dreamtalk} further achieve more vivid results. Recently, a major shift occurred with the introduction of text-to-image pre-trained models. EMO~\cite{emo}, Hallo~\cite{hallo} and other similar frameworks~\cite{aniportrait,vexpress} built on the foundation of pre-trained image diffusion models~\cite{latentdiffusion}, achieving high-fidelity talking head video generation. Traditional audio-driven methods simply based on a data-driven approach, lack optional control on expression styles. The large-scale network and widely-used auto-regressive inference strategy~\cite{dreamtalk,hallo,vasa,gaia} also constrain their ability to generate stable long-time videos.

\subsection{Controllable Talking Head Generation}
Controlling the expression style in talking head video generation has long been a compelling challenge. Early methods~\cite{aevp,egtfg,emmn,emote,eat,space} model expressions in discrete emotion states, while recent methods~\cite{styletalk,dreamtalk,pdfgc,anitalker} focus on transferring the expressions from a reference video to the generated video. Extracting decoupled representations of expressions is crucial for emotion transferring. Earlier approaches~\cite{styletalk,dreamtalk} use 3DMM coefficients~\cite{3dmm,3dmmsurvey} from reference videos, but this led to identity leakage issues, as the 3DMM coefficients encode not only expression information but also the speaker's facial structure information. PD-FGC~\cite{pdfgc} and Anitalker~\cite{anitalker} employ contrastive learning approaches to acquire expression-related latent and realize expression driven with minor identity leakage.

In practical applications, emotion control information can originate from many other sources~\cite{ersurvey,mersurvey}. In our approach, we derive a unified emotional control latent from various optional sources of emotion information and enable emotion control based on the emotion control latent.

\subsection{Video Diffusion Models}
The pioneering work on text-to-video diffusion is Video Diffusion Models (VDM)~\cite{videodiffusionmodels}. ImagenVideo~\cite{imagenvideo} enhances VDM with cascaded diffusion models. Make-A-Video ~\cite{makeavideo} and MagicVideo~\cite{magicvideo} then extend these concepts to enable seamless text-to-video transformations. AnimateDiff~\cite{animatediff} utilizes a motion module to realize the conversion from text-to-image to text-to-video. Stable Video Diffusion (SVD)~\cite{stablevideodiffusion} implements innovative training strategies to generate high-fidelity videos. In our research, we use diffusion models in both expression motion latent generating and talking head rendering under facial motion control conditions.

%% file: sec/3_finalcopy.tex
\begin{figure*}[!ht]
  \centering
  \includegraphics[width=1.0\linewidth]{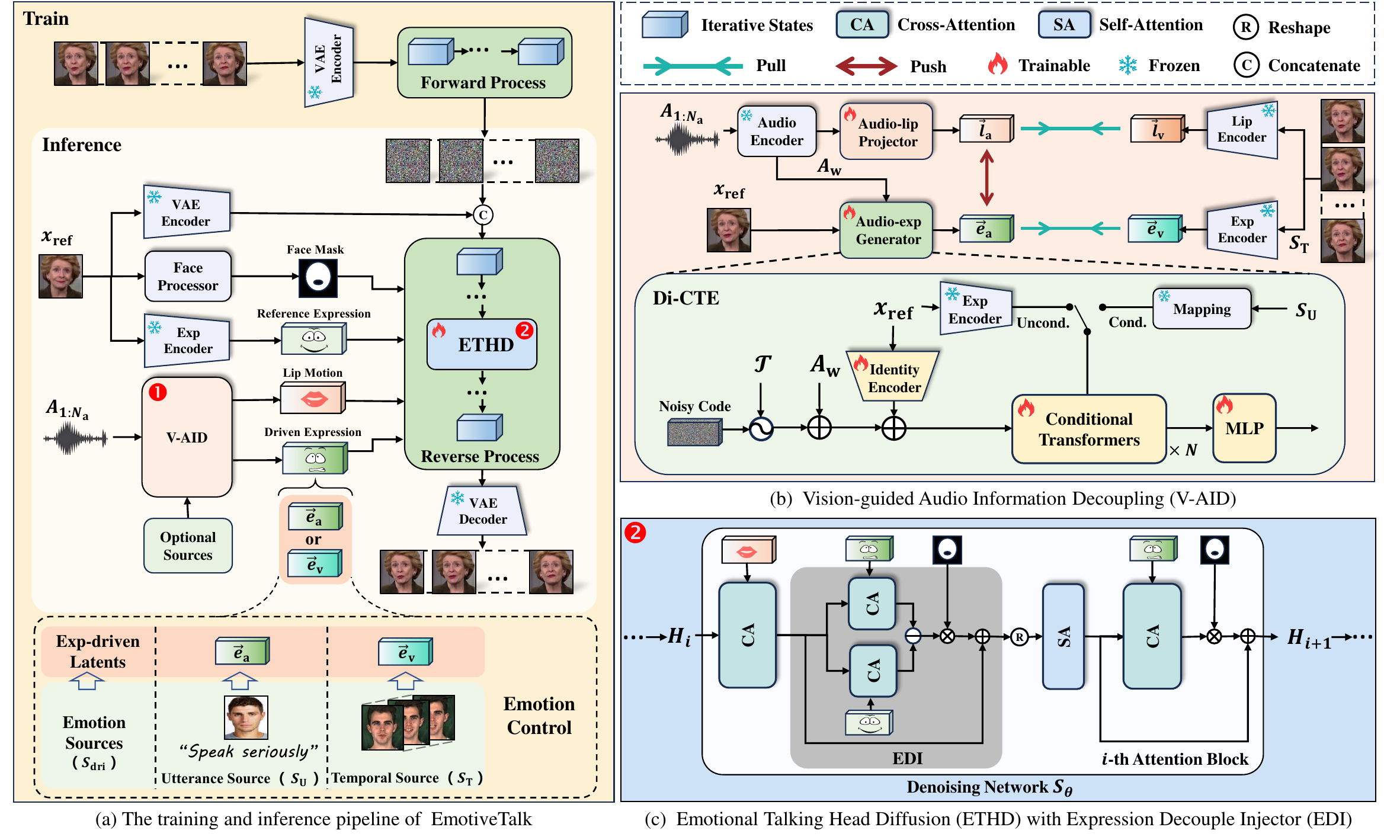}
  \caption{The framework of EmotiveTalk. During the training process, the Vision-guided Audio Information Decouple (V-AID) module with Diffusion-based Co-speech Temporal Expansion (Di-CTE) expression generator in (b) is firstly trained to provide lip-related and expression-related representation from audio. Then the Emotional Talking Head Diffusion (ETHD) framework with Expression Decouple Injector (EDI) in (c) is trained with reference portrait condition and facial motion conditions to reconstruct the target frames, including lip-related and emotion-driven representation randomly chosen between $\mathit{\boldsymbol{\Vec{e}}}_\text{a}$ and $\mathit{\boldsymbol{\Vec{e}}}_\text{v}$ from V-AID module. During the inference process, EmotiveTalk takes portrait and speech audio as input, supplemented with optional emotion source $\mathit{\boldsymbol{S}_\text{dri}}$ to achieve emotion control.}
  \label{fig:overall}
\end{figure*}

\section{Method}

As shown in Fig.~\ref{fig:overall}, the structure of the EmotiveTalk is divided into two main parts: (1) the Vision-guided Audio Information Decouple (V-AID) with Diffusion-based Co-speech Temporal Expansion (Di-CTE) module; (2) the Emotional Talking Head Diffusion (ETHD) framework with Expression Decoupling Injector (EDI) module.

\subsection{Preliminary}
\noindent\textbf{Task Definition.} The task of controllable talking head generation involves creating a vivid talking head video from two inputs: a static single-person portrait $\mathit{\boldsymbol{x}_\text{ref}}$, and a driven speech sequence $\mathit{\boldsymbol{A}} \in \mathbb{R}^{N_a}$. Besides, emotion sources $\mathit{\boldsymbol{S}_\text{dri}}$ can also utilized as optional input to realize better controllability on emotion. When optional $\mathit{\boldsymbol{S}_\text{dri}}$ is not provided, our method aims to generate expression-related representation solely by the speech input and the portrait $\mathit{\boldsymbol{x}_\text{ref}}$. The output is the generated video frames $ \hat{\mathit{\boldsymbol{X}}}_{1:N} = \{ \hat{\mathit{\boldsymbol{x}}}_{0}, ..., \hat{\mathit{\boldsymbol{x}}}_{N} \} $.

\noindent\textbf{Diffusion Models.} 
% Diffusion models belong to the class of generative models that generate samples conforming to a given data distribution by progressively denoising Gaussian noise~\cite{ddpm}. 
Let $\boldsymbol{X}_{(0)}$ represent video latents sampled from a given distribution $q(\boldsymbol{X}_{(0)})$. In the forward diffusion process, Gaussian noise is progressively added to $\boldsymbol{X}_0$, gradually diffusing towards a distribution resembling $\mathcal{N}(\mathbf{0}, \mathit{\boldsymbol{I}})$. This process forms a fixed Markov chain \cite{ddpm,ddim}:
\begin{equation}
    q(\boldsymbol{X}_{(t)}|\boldsymbol{X}_{(t-1)}) = \mathcal{N}(\boldsymbol{X}_{(t)}; \sqrt{1-\beta_t} \, \boldsymbol{X}_{(t-1)}, \beta_t\boldsymbol{I})
\end{equation}

\noindent where $ \{\beta_t\}_{t=1}^T $ are known constants. Notably, the marginal distribution at any time can directly derive from $ \boldsymbol{X}_{(0)} $ as:
\begin{equation}
q(\boldsymbol{X}_{(t)} | \boldsymbol{X}_{(0)} = \mathcal{N}(\boldsymbol{X}_{(t)}; \sqrt{\bar{\alpha}_t} \, \boldsymbol{X}_{(0)}, (1-\bar{\alpha}_t)\boldsymbol{I})
\end{equation}
where $ \bar{\alpha}_t = \prod_{i=1}^t \alpha_i $ and $ \alpha_t = 1 - \beta_t $. The reverse process gradually recovers the original video latent from the noisy latent $ \boldsymbol{X}_{(T)} \sim \mathcal{N}(\mathbf{0}, \mathit{\boldsymbol{I}})$, achieving by training a network to predict the posterior distribution $ p_\theta(\boldsymbol{X}_{(t-1)} | \boldsymbol{X}_{(t)}, \boldsymbol{c}) $ under condition set $\boldsymbol{c}$. To learn $ p_\theta(\boldsymbol{X}_{(t-1)} | \boldsymbol{X}_{(t)}) $. The model is trained using the following loss function:
\begin{equation}
\label{eq:core_dm_loss}
L = \mathbb{E}_{t, \boldsymbol{X}_0, \epsilon, \boldsymbol{c}} [\| \epsilon - \epsilon_\theta(\boldsymbol{X}_{(t)}, t, \boldsymbol{c}) \|^2]
\end{equation}
 We use diffusion strategy for expression-related representation generation in Di-CTE and video latent generation.

\subsection{Vision-guided Audio Information Decouple}
\label{sec:V-AID}
% Previous studies have shown that features with relatively strong temporal dynamics in speech signals, such as phoneme and prosody information, are highly related to lip movements in video, while features with weaker temporal dynamics, such as emotion information, are more closely associated with facial expressions. 
Speech is rich in plentiful coupled information, previous methods focused on decoupling speech information in the audio space~\cite{audiodecouple, instructtts}. However, the representations obtained through these approaches are generally not well-suited for talking head generation, due to the inherent disparity between the audio and facial motion representations. We propose that facial motion information in the vision space can guide the decoupling of coupled speech information due to the correlation between speech information and different facial motions and also facilitate the generation of aligned facial motion related representations from audio. Based on this, we designed a Vision-guided Audio Information Decoupling (V-AID) module. This module takes audio sequence $\mathit{\boldsymbol{A}}$ and reference portrait $\boldsymbol{x}_\text{ref}$ as input. The audio stream first passes through a pre-trained Wav2Vec audio encoder~\cite{wav2vec}, followed by the trainable audio-to-lip projector and audio-to-expression generator to obtain lip and expression-related latents. The two modules are trained under the supervision of lip and expression representations of vision space, elaborated in the supplementary material. 

\noindent\textbf{Audio-lip Contrastive Learning.}
We leverage the latent representation of lip motions in vision space to guide the audio-to-lip mapping, thereby achieving alignment between the audio and lip motion representations. Specifically, we use a pre-trained lip encoder to extract decoupled lip-related latents $ \mathit{\boldsymbol{\Vec{l}}}_\text{v} = \{ \mathit{{l}}_{1}, ..., \mathit{{l}}_{N} \} $ from videos paired with audio. The audio stream is processed through an audio-to-lip projector with a Perceiver Transformer~\cite{perceiver} architecture detailed in the supplementary material to generate lip-related latents $ \mathit{\boldsymbol{\Vec{l}}}_\text{a} = \{\mathit{{\hat{l}}}_{1}, ..., \mathit{{\hat{l}}}_{N} \} $. The infoNCE~\cite{infonce} contrastive loss function is utilized to optimize the lower bound of mutual information (MI) between $ \mathit{\boldsymbol{\Vec{l}}}_\text{a}$ and $ \mathit{\boldsymbol{\Vec{l}}}_\text{v}$ to maximize MI between frame-level lip movements and the corresponding driving speech signal, where $(\mathit{{\hat{l}}}_{i}, \mathit{{l}}_{i})$ denotes a positive pair and $(\mathit{{\hat{l}}}_{i}, \mathit{{l}}_{j}) $ denotes negative pairs. The loss function is formulated as follows, with $\mathrm{sim}(\cdot)$ represents cosine similarity:  
\begin{equation}
  \mathcal{L}_{\text{lipc}} =-\frac{1}{N} \sum_{i = 1}^{N} \mathrm{log} \left (\frac{\mathrm{exp}\left ((\frac{\mathrm{sim}(\hat{l}_{i} , l_{i})}{\tau}\right)}{\sum_{j = 1}^{N} \mathrm{exp}\left(\frac{\mathrm{sim}(\hat{l}_{i} , l_{i})}{\tau}\right)}\right) 
\end{equation}
Furthermore, we also supplement the contrastive learning loss with Mean Squared Error (MSE) loss to synchronize both the motion and morphological information between $\mathit{\boldsymbol{\Vec{l}}}_\text{a}$ and $\mathit{\boldsymbol{\Vec{l}}}_\text{v}$. The loss function is as follows:
\begin{equation}
  \mathcal{L}_{\text{lipm}} =\frac{1}{N} \sum_{i = 1}^{N} ||l_{i} - \hat{l}_{i}||^2 
\end{equation}
The final training loss function is the combination of two losses, as follows:

\begin{equation}
  \mathcal{L}_\text{lip} = \alpha \mathcal{L}_{\text{lipc}} + \beta \mathcal{L}_{\text{lipm}}
  \label{eq:a2l loss}
\end{equation}

% \begin{equation}  too long so ugly
%   \mathcal{L}_{\mathrm{lip_c}} =\frac{1}{N} \sum_{i = 1}^{N}( - \mathrm{log} (\frac{\mathrm{exp}(\frac{\hat{l}_{i} \cdot l_{i}}{\tau})}{\sum_{j = 1}^{N} \mathrm{exp}(\frac{\hat{l}_{i} \cdot l_{i}}{\tau})}) + (l_{i} - \hat{l}_{i})^2)
% \end{equation}

\noindent\textbf{Di-CTE for Audio-to-expression Generation.} 
We utilize representations of facial expressions from the vision space to guide the alignment of audio-based emotion information with facial expressions. Generally, speech and facial expressions are not strictly correlated on a one-to-one basis, the same speech can correspond to different but plausible facial expressions. To address this, we propose a Diffusion-based Co-speech Temporal Expansion (Di-CTE) module to generate frame-level expression-related latent $\mathit{\boldsymbol{\Vec{e}}}_\text{a}$ from initial expression under speech constraints, leveraging the advantages of diffusion models in terms of generative diversity. We leverage a pre-trained expression encoder to extract decoupled expression latent $\mathit{\boldsymbol{\Vec{e}}}_\text{v}$ from ground-truth video as vision supervision. Di-CTE inputs consist of a reference frame ($\mathit{\boldsymbol{x}_\text{ref}}$) from the ground truth video serving as speaker identity and speech embedding $\mathit{\boldsymbol{A}}_\text{w}$ to provide temporal emotion information. During training, the emotion condition $\mathit{\boldsymbol{e}_\text{cond}}$ is provided by the first frame of the ground-truth video, and the output is expression-related latent $\mathit{\boldsymbol{\Vec{e}}}_\text{a}$ sync with the speech. The denoising loss of network $S_{\theta}$ is defined as follows, where t denotes the DDPM step:
\begin{equation}
  \mathcal{L}_\text{exp} = ||\mathit{\boldsymbol{\Vec{e}}}_\text{v} - S_{\theta}(\mathit{\boldsymbol{\Vec{e}}}_{\text{a}(t)}, t, \mathit{\boldsymbol{x}_\text{ref}}, \mathit{\boldsymbol{A}}_\text{w}, \mathit{\boldsymbol{e}_\text{cond}})||^2 %_2^2
  \label{eq:Di-CTE loss}
\end{equation}

\noindent\textbf{Mutual Information Constraint.}
Finally, to decouple lip-related and expression-related information and mitigate their mutual interference, we introduced a mutual information (MI) constraint during the joint training of the audio-to-lip and audio-to-expression modules. Specifically, we employ CLUB~\cite{club} to optimize the upper bound of MI between the lip-related latent $\mathit{\boldsymbol{\Vec{l}}}_\text{a}$ from the audio-to-lip module and the expression-related latent $\mathit{\boldsymbol{\Vec{e}}}_\text{a}$ from the audio-to-expression module. The total loss function is as follows:
\begin{equation}
  \mathcal{L}_\text{V-AID} = \mathcal{L}_\text{lip} + \mathcal{L}_\text{exp} + \rm{CLUB}\{\mathit{\boldsymbol{\Vec{l}}}_\text{a}, \mathit{\boldsymbol{\Vec{e}}}_\text{a}\}
  \label{eq:joint loss}
\end{equation}
By minimizing the MI between $\mathit{\boldsymbol{\Vec{l}}}_\text{a}$ and $\mathit{\boldsymbol{\Vec{e}}}_\text{a}$, we achieve a separation of the two representation spaces. 

\subsection{Emotional Talking Head Diffusion}
\label{sec:ETHD}
In this subsection, we present a diffusion-based framework for generating emotional talking heads. Before ETHD, the driving audio is processed through V-AID to obtain lip-related and expression-related latents, and the portrait is projected into latent space via temporal Variational Autoencoder (VAE) and concatenated with input noise along the channel dimension, shown in Fig.~\ref{fig:overall}. ETHD outputs a sequence of frame latents synchronized with the speech. 

\noindent\textbf{Backbone Network.}
Our backbone network leverages a 3D-Unet architecture with the spatial-temporal separable attention mechanism~\cite{stablevideodiffusion}. The spatial attention module comprises two blocks. Firstly, the lip-related latent $\mathit{\boldsymbol{\Vec{l}}}_{\text{a}}$ is injected through spatial cross attention. Then, an Expression Decoupling Injector (EDI) module, articulated late in Section \ref{sec:EDI},  is employed to integrate expression-driven latent $\mathit{\boldsymbol{\Vec{e}}}_{\text{dri}}$ ($\mathit{\boldsymbol{\Vec{e}}}_{\text{dri}}=\mathit{\boldsymbol{\Vec{e}}}_{\text{a}}$ for audio-driven task and $\mathit{\boldsymbol{\Vec{e}}}_{\text{dri}}=\mathit{\boldsymbol{\Vec{e}}}_{\text{v}}$ for video-driven task). Analogously, the temporal attention module also encompasses two components: a temporal self-attention mechanism and a temporal cross-attention module. The temporal cross-attention module engages in cross-attention with expression-driven latent to learn subtle temporal variations in emotional expression. The output latents are then processed through a temporal VAE decoder to obtain the generated motion frames.

\noindent\textbf{Expression Decoupling Injector.}
\label{sec:EDI}
In talking head generation, the inherent expression information in the reference portrait usually constrains the generation of the target expression, leading to sub-optimal expressive results. To address this, we propose an Expression Decoupling Injection (EDI) module to achieve emotional expressions by automatically decoupling the expression information from reference portraits while integrating the expression-driven information, which consists of two parallel attention branches. One branch computes the attention between the hidden states $\mathit{\boldsymbol{H}}_{i} \in \mathbb{R}^{f \times h \times w \times c}$ ($f$ is the number of processed frames, $h$ and $w$ is the height and width of hidden states, $c$ is the number of channels) and the expression embeddings $\mathit{\boldsymbol{\Vec{e}}}_{\text{ref}}$ of the reference portrait while the other branch computes the attention between the hidden states $\mathit{\boldsymbol{H}}_{i}$ and the expression-driven representation $\mathit{\boldsymbol{\Vec{e}}}_{\text{dri}}$. By subtracting these two cross-attention outputs, we achieve the transition of facial expressions in the generated video from the expression of the reference image to the driving expression, as shown in the following equation:
\begin{equation}
  \boldsymbol{Attn}_{i}\!=\!\rm{CrossAttn}(\mathit{\boldsymbol{H}}_{\mathit{i}}, \mathit{\boldsymbol{\Vec{e}}}_{dri})\!-\!\rm{CrossAttn}(\mathit{\boldsymbol{H}}_{\mathit{i}}, \mathit{\boldsymbol{\Vec{e}}}_{\text{ref}})
\end{equation}

Moreover, to enforce the expression-related latent act only on the facial region without affecting the lip region generation, we apply an attention mask similar to Hallo~\cite{hallo} to the resulting attention value. Specifically, we use the off-the-shelf toolbox OpenFace~\cite{openface} to predict landmarks from portrait $\mathit{\boldsymbol{x}_{\text{ref}}}$ and calculate binary bounding box masks $\mathbf{M}_{\text{lip}}, \mathbf{M}_{\text{face}} \in \{0,1\}^{h \times w}$ which indicate the inner of lip region and face region. Then, the output of the EDI block is formulated based on bounding box masks, as follows:
\begin{equation}
  \mathit{\boldsymbol{H}}_{i}^{\text{spa}} = \mathit{\boldsymbol{H}}_{i} + \mathit{\boldsymbol{Attn}}_i \odot (1-\mathbf{M}_{\text{lip}}) \odot \mathbf{M}_{\text{face}}
\end{equation}

\noindent\textbf{Expression Temporal Cross-attention.}
To implement better modeling of the time-variance of facial expressions, we introduce a temporal cross-attention module. Specifically, we squeeze the spatial dimensions of the hidden states $\mathit{\boldsymbol{H}}_{i}^{\text{spa}}$ to $\mathit{\boldsymbol{H}}_{i}^{\text{tem}} \in \mathbb{R}^{(h \times w) \times f \times c}$ and compute the cross-attention between $\mathit{\boldsymbol{H}}_{i}^{\text{tem}}$ and the expression-driven latent $\mathit{\boldsymbol{\Vec{e}}}_{\text{dri}}$. This makes the model more sensitive to the temporal correlations of emotional information. Additionally, the same bounding box masks are utilized to constrain the sensible area of attention calculation. 

\subsection{Training and Inference}
\label{sec:training and inference}

\noindent\textbf{Training.}
The V-AID module in Sec.~\ref{sec:V-AID} is first pre-trained to generate decoupled lip-related representation $\mathit{\boldsymbol{\Vec{l}}}_\text{a}$ and emotion-related representation $\mathit{\boldsymbol{\Vec{e}}}_\text{a}$ from driven audio window $A_{\text{w}}$ and then remain frozen while training the ETHD backbone.

Subsequently, we train the ETHD backbone by sampling tuples $(\boldsymbol{X}, \mathit{\boldsymbol{x}_{\text{ref}}}, t ,\mathit{\boldsymbol{\Vec{l}}}_\text{a}, \mathit{\boldsymbol{\Vec{e}}}_{\text{ref}}, \mathit{\boldsymbol{\Vec{e}}}_{\text{dri}})$, $\mathit{\boldsymbol{\Vec{e}}}_{\text{dri}}$ is random choice in video expression-related representation $\mathit{\boldsymbol{\Vec{e}}}_\text{v}$ and the generated expression-related representation $\mathit{\boldsymbol{\Vec{e}}}_\text{a}$. The total denoising loss function is formulated as:
\begin{equation}
\mathcal{L}_{\text{de}} = ||\boldsymbol{X}_{(0)} - S_{\theta}(\boldsymbol{X}_{(t)}, \boldsymbol{x}_{\text{ref}}, t, \mathit{\boldsymbol{\Vec{l}}}_\text{a}, \mathit{\boldsymbol{\Vec{e}}}_{\text{ref}}, \mathit{\boldsymbol{\Vec{e}}}_{\text{dri}})||^2 %_2^2
  \label{eq:Denoise loss}
\end{equation}

\noindent\textbf{Inference.} In the inference phase, we employ a non-autoregressive inference method to avoid the accumulation of error. Specifically, when performing long-time generation, we sample a Gaussian-like noisy latent and divide the total duration into several overlapping clips with a defined window size. We utilize DDIM~\cite{ddim} sampler for ETHD to denoise each clip sequentially per step, then we assign a weighting strategy the same as Mimicmotion~\cite{mimicmotion} to assign higher fusion weights for frame latents closer to the center of each clip. Repeat this process iteratively to obtain the clean frame latent. This approach allows us to perform inference of arbitrary lengths without error accumulation.

% \noindent\textbf{Inference.} In the inference phase, we employ a non-autoregressive inference method to avoid the accumulation of error. Specifically, when performing long-time generation, we sample a Gaussian-like noisy latent 
% $X_(T)\in$

% divide the total duration into several overlapping clips with a defined window size. We utilize DDIM~\cite{ddim} method for denoising the noise of each clip sequentially per step, then we assign a weighting stategy the same as Mimicmotion~\cite{mimicmotion} to assign higher fusion weights for frames closer to the center of each clip. Then the weighted This approach allows us to perform inference of arbitrary lengths without the influence of error accumulation.

\subsection{Multi-source Emotion Control}
\label{sec:MEC}
To flexibly control emotional expression in generated video based on control sources, we designed the Multi-source Emotion Control (MEC) pipeline. MEC introduces time-varying facial expressions to the generated video based on optional temporal or utterance sources.

\noindent\textbf{Temporal Sources Emotion Control.} External expression-driven videos are treated as temporal sources, denoted as $\mathit{\boldsymbol{S}_{\text{T}}}$, due to their rich temporal variations in expression. We directly apply the pre-trained expression encoder to extract the expression-driven latent $\mathit{\boldsymbol{\Vec{e}}}_{\text{v}}$, as detailed in Section~\ref{sec:ETHD}. The final emotive video is rendered using exp-driven latent $\mathit{\boldsymbol{\Vec{e}}}_{\text{v}}$, and lip-related latent $\mathit{\boldsymbol{\Vec{l}}}_{\text{a}}$, derived from Section~\ref{sec:V-AID}.

\noindent\textbf{Utterance Sources Emotion Control.} To improve the temporal dynamism and better alignment with the driving speech of generated expressions based on utterance sources $\mathit{\boldsymbol{S}_{\text{U}}}$ that only 
provide general emotional information $\mathit{\boldsymbol{e}}_{\text{cond}}$, we use the Di-CTE module (Section~\ref{sec:V-AID}) to generate frame-level expression-driven latent $\mathit{\boldsymbol{\Vec{e}}}_{\text{a}}$ from $\mathit{\boldsymbol{e}}_{\text{cond}}$. Specifically, for expression-driven images of different people ($\mathit{\boldsymbol{x}}_{\text{dri}}$), we map the image to the emotion condition latent space $\mathit{\boldsymbol{e}}_{\text{cond}}$ using pre-trained expression encoder (Section~\ref{sec:V-AID}). For cross-modality control sources like $\mathit{\boldsymbol{t}}_{\text{dri}}$, we apply a cross-modality mapping to align with $\mathit{\boldsymbol{e}}_{\text{cond}}$, detailed in the supplementary material. The final emotive video is rendered using lip-related latent $\mathit{\boldsymbol{\Vec{l}}}_{\text{a}}$ and expression-driven latent $\mathit{\boldsymbol{\Vec{e}}}_{\text{a}}$ via our diffusion backbone, as detailed in Section~\ref{sec:ETHD}.

\begin{table*}[]
\renewcommand{\arraystretch}{1.2}
\centering
\tabcolsep=0.02\linewidth
\begin{tabular}{c|cccccc}
\multirow{2}{*}{Methods} & \multicolumn{6}{c}{HDTF / MEAD}                                                            \\
                         & Driven & FID ($\downarrow$)         & FVD ($\downarrow$)          & Sync-C ($\uparrow$)   & Sync-D ($\downarrow$)     & E-FID ($\downarrow$)    \\ \hline
SadTalker~\cite{sadtalker}                & A  & 22.34 / \textbf{36.88}      & 589.63 / \textbf{132.27} & 7.75 / 6.46 & 7.36 / 8.07   & 0.66 / 1.14 \\
Anitalker~\cite{anitalker}                & A   & 51.66 / 68.01  & 583.70 / 941.49 & 7.73 / 6.76 & 7.43 / 7.64   & 1.11 / 1.11 \\
Aniportrait~\cite{aniportrait}              & A   & 17.71 / 42.43  & 676.30 / 379.08 & 3.75 / 2.30 & 10.63 / 12.38 & 1.21 / 2.69 \\
Hallo~\cite{hallo}                    & A  & 17.15 / 52.07          & 276.31 / 210.56 & 7.99 / \textbf{7.45} & 7.50 / 7.47   & 0.65 / 0.60    \\
\rowcolor{green!8}
Ours & A & \textbf{16.64} / 53.21          & \textbf{140.96} / 207.67 & \textbf{8.24} / 6.82 & \textbf{7.09} / \textbf{7.43} & \textbf{0.54} / \textbf{0.57} \\ \hline
PD-FGC~\cite{pdfgc}                   & A+V   & 67.97 / 121.46      & 464.90 / 353.75      & 7.30 / 5.15    & 7.72 / 8.77      & 0.74 / 1.92    \\
StylkTalk~\cite{styletalk}                & A+V     & 29.65 / 118.48 & 184.60 / 197.18 & 4.34 / 3.86 & 10.35 / 10.74 & 0.42 / 0.56 \\
DreamTalk~\cite{dreamtalk}                & A+V   & 29.37 / 105.92 & 263.78 / 204.48 & 6.80 / 5.64 & 8.03 / 8.69   & 0.55 / 0.87 \\
\rowcolor{green!8}
Ours & A+V & \textbf{16.09} / \textbf{50.84} & {\textbf{120.70} / \textbf{153.71}}    & \textbf{8.41} / \textbf{6.79}       & \textbf{7.11} / \textbf{7.58}       & {\textbf{0.34} / \textbf{0.40}} \\ \hline
Ground Truth             & A+V   & -         & -          & 8.63 / 7.30       & 6.75 / 8.31         & -      
\end{tabular}
\caption{Overall comparisons on HDTF and MEAD. "A" denotes audio-driven and "A+V" denotes audio-video driven. ``$\uparrow$'' indicates better performance with higher values, while ``$\downarrow$'' indicates better performance with lower values.}
\label{audionly_results}
\end{table*}

\section{Experiment}
\label{sec:experiments}

\subsection{Experimental Setup} 
\label{sec:setup}
\noindent\textbf{Implementation Details}.
Experiments encompassing both training and inference were carried out on open-source datasets HDTF~\cite{hdtf} and MEAD~\cite{mead}, which  consist of talking individuals videos of diverse genders, ages, and ethnicities. We utilize a two-stage training strategy, firstly, we trained the V-AID module with a learning rate of 1e-4. In the second stage, the audio-to-video diffusion backbone was trained while the pre-trained V-AID modules remained frozen in training. Notably, thanks to the efficient design of our model, we can conduct high-resolution and long-time video training. We conduct a training configuration of the resolution of $512\times512$ and 120 frames. The learning rate is set to 1e-5 with a batch size of 1. Our backbone also supports up to $1024\times1024$ training, and experiments on other configurations are detailed in the supplementary material.

During the inference, we use the sampling algorithm of DDIM~\cite{ddim} to generate the video clip for 25 steps, the inference window size is as same as the training frame number and the overlap is set to 1/5 of the window size.

\noindent\textbf{Evaluation Metrics}.
The proposed framework has been evaluated with several quantitative metrics including Fréchet Inception Distance (FID)~\cite{fid}, Fréchet Video Distance (FVD) ~\cite{fvd,stylegan_v}, Synchronization-C (Sync-C)~\cite{syncnet}, Synchronization-D (Sync-D)~\cite{syncnet} and E-FID~\cite{emo}. Specifically, FID and FVD conduct the image-level and frame-level measurement of the quality of the generated frames and the similarity between generated and ground-truth frames, with lower values indicating better performance. The SyncNet scores assess the lip synchronization quality, with higher Sync-C and lower Sync-D scores indicating better alignment with the driven speech signal. Additionally, to evaluate the expressiveness of the facial expressions in the generated videos, we also utilize the Expression-FID (E-FID) metric introduced in EMO~\cite{emo} to quantitatively measure the expression divergence between the synthesized videos and gound-truth videos.

\noindent\textbf{Baselines}.
We conducted a comparative analysis of our proposed method against several open-source implementations, including audio-driven strategies including SadTalker~\cite{sadtalker}, AniPortrait~\cite{aniportrait}, Anitalker~\cite{anitalker} and Hallo~\cite{hallo}, and audio-video driven strategies including PDFGC~\cite{pdfgc}, styletalk~\cite{styletalk} and dreamtalk~\cite{dreamtalk}. For audio-driven comparison, our framework derives lip-related and expression-driven latents solely from the audio and reference portrait input. As for audio-video driven, the expression-driven latent is derived from the paired video.

\begin{figure*}[ht]
    \centering
    \includegraphics[width=1.0\linewidth]{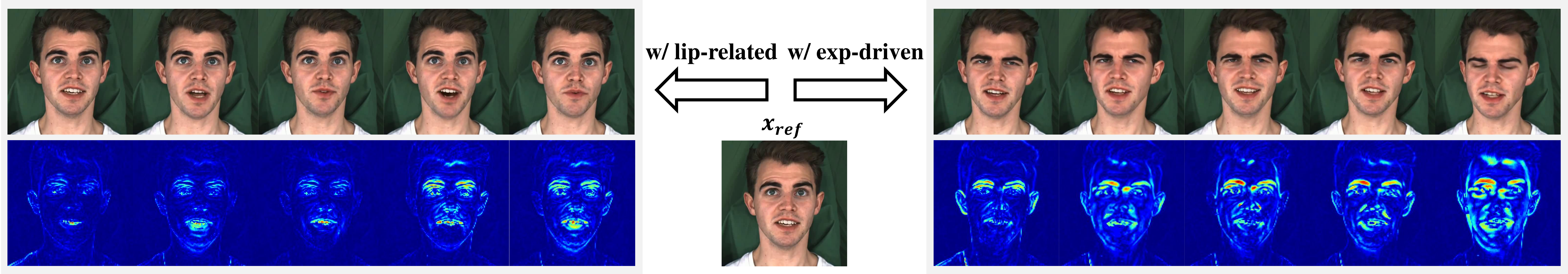}
    \caption{The visualize results of generated frames and difference heatmap with the reference portrait based on lip-related and expression-driven representations driven separately.}
    \label{fig:V-AID_heatmap}
\end{figure*}

\subsection{Overall Evaluation}
\label{sec:overallevaluation}
Tab.~\ref{audionly_results} shows the results of the comprehensive comparison with other methods. Overall, methods based on pre-trained diffusion models like Hallo or Aniportrait achieve optimal FID metrics, confirming the potential of diffusion models in generating high-fidelity videos. Also, video-driven approaches perform better on the E-FID metric benefiting from the inclusion of expression cues derived from video. Our method outperforms previous methods in both audio-driven and video-driven tasks across most metrics, especially on E-FID and SyncNet metrics, highlighting its superior capabilities of generating high-fidelity and vivid videos.

\subsection{Ablation Study}
To analyze the contributions of our designs, we conduct ablation studies on our main modules.

\noindent\textbf{Effectiveness of V-AID.}
We conduct an ablation study with three variants: (1) driven only by original audio embedding without V-AID (no decouple); (2) driven without lip-related latents (w/o lip-related); (3) driven without expression-driven latents (w/o exp-driven). Our full model is denoted as (V-AID). The experiment is carried out in the test subset of HDTF. Shown in Tab~\ref{ablation_V-AID}, the results indicate that using V-AID shows improvements across all three metrics compared to direct injection without decoupling, with notable gains in the Sync-C and E-FID metrics. Additionally, we observe a significant drop in Sync-C when lip-related latents are removed, and a substantial degradation in E-FID when expression-driven latents are excluded. This supports the different roles that the two representations play in driving lip movement and facial expressions. Furthermore, we observe that FID achieves the best performance without expression-driven, this is due to the higher similarity between generated frames and reference images when expression-driven latents are excluded, further confirmed in subsequent experiments.

\begin{table}[]
\centering
\begin{tabular}{@{}cccc@{}}
Methods                   & FID ($\downarrow$) & Sync-C ($\uparrow$) & E-FID ($\downarrow$) \\ \midrule
\rowcolor{green!8}
V-AID        &     16.64    &     \textbf{8.24}      &   \textbf{0.54}    \\ 
w/o lip-related &   16.02     &      0.65     &   0.57    \\ 
w/o exp-driven &   \textbf{14.86}     &    8.04      &  1.23     \\ 
no decouple     &   16.98    &    7.72       &     0.66
\end{tabular}
\caption{Ablation comparison on V-AID on HDTF dataset.}
\label{ablation_V-AID}
\end{table}

\noindent\textbf{Effectiveness of Decoupled Representations.}
To evaluate the decoupling ability of two representations, we utilized the lip-related and expression-driven latents from V-AID to generate video separately and visualize the results. Shown in Fig.~\ref{fig:V-AID_heatmap}, the visualized results indicate that the main movement occurs at the lip region, with higher heat values concentrated around the lip region. In contrast, the generated frames driven by the expression-driven latents exhibit substantial changes in facial expressions compared to the reference portrait, with higher heat values distributed across the entire facial area, particularly in the eye region associated with angry emotion. The results demonstrate the effectiveness of decoupled lip and expression representations in controlling facial motions separately.

% \begin{figure*}[ht]
%     \centering
%     \includegraphics[width=1.0\linewidth]{fig5.eps}
%     \caption{The visualize results of generated frames and difference heatmap with the reference portrait with E-CTE module activated and without E-CTE module activated.}
%     \label{fig:E-CTE_heatmap}
% \end{figure*}

% \noindent\textbf{Ablation study on MEC.} 

% To validate the superiority of our proposed E-CTE module in generating dynamic expressions over time, we employed the same expression-driven image and conducted inference using two configurations: with the E-CTE module activated (w. E-CTE) and without the E-CTE module activated (w/o E-CTE) for emotion style transfer separately. As shown in Fig.~\ref{fig:E-CTE_heatmap}, with and without the activation of E-CTE module both yield satisfactory expression transfer results. However, the facial expressions in inference results without the E-CTE module activated show minimal temporal variation, remaining largely consistent with the expression of the driving image. This is reflected in the difference heatmap, where areas of high heat values remain relatively stable throughout the duration. In contrast, the facial expressions in inference results with the E-CTE module activated exhibit significant temporal variation, resulting in more dynamic and expressive outputs. The differece heatmap in this case shows pronounced changes in the regions of high heat values over time. The visualization results demonstrate the effectiveness of our E-CTE module in generating vivid, time-varying expressions. 

\begin{figure}[t]
    \centering
    \includegraphics[width=1.0\linewidth]{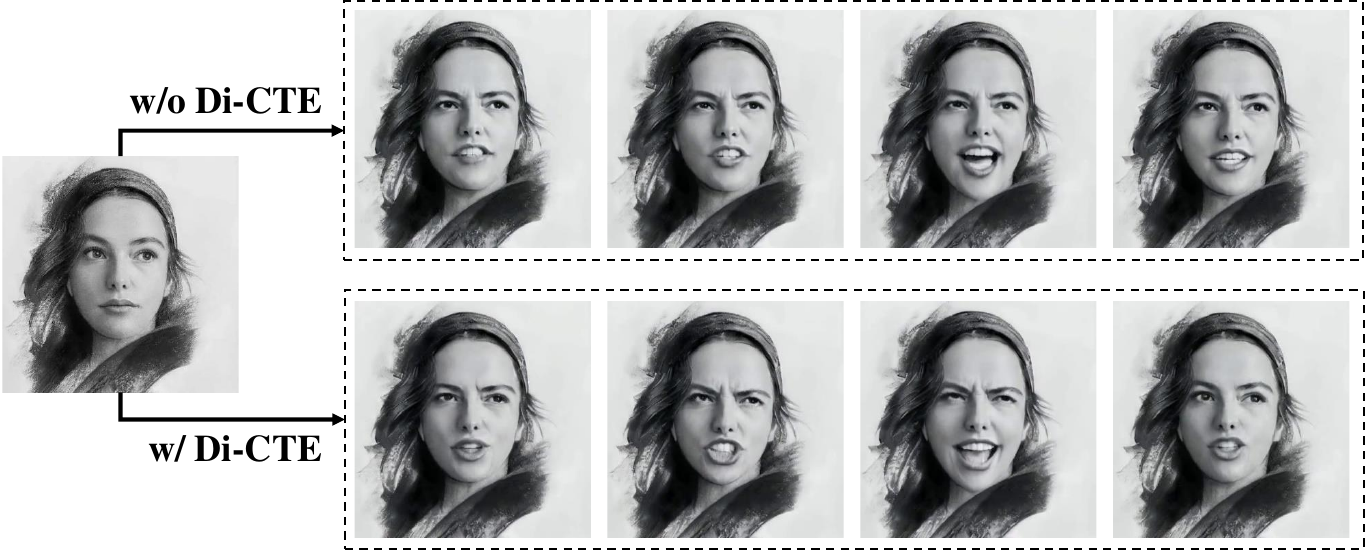}
    \caption{Results on expression generation w/ or w/o Di-CTE.}
    \label{fig:DiCTE}
\end{figure}

\noindent\textbf{Effectiveness of Di-CTE.} To validate the superiority of our proposed Di-CTE module in expanding utterance driven sources to generate time-variance expressions, we employed a single expression-driven image and conducted inference using two configurations: with the Di-CTE module (w/ Di-CTE) and without the Di-CTE module (w/o Di-CTE). The facial expressions in inference results w/o Di-CTE module activated show minimal temporal variation in Fig.~\ref{fig:DiCTE}, while more expressive and vivid results are achieved by Di-CTE activated, demonstrating its effectiveness.

\noindent\textbf{Effectiveness on Long-time Generation.}
To validate the stability of long-time generation, we conducted studies on generating with varying lengths by audio-driven. We employed four different test configurations, ranging from short to long duration, and evaluated identity consistency, lip-sync accuracy, and expression alignment across varying generation durations. The results are presented in Tab~\ref{ablation_longtime}.

% \begin{table}[]
% \caption{Ablation comparison on varying-time generation.}
% \label{ablation_longtime}
% \centering
% \tabcolsep=0.028\linewidth
% \begin{tabular}{ccccc}
% \hline
% \diagbox{Length}{Metric} & FID ($\downarrow$) & Sync-C ($\uparrow$) & E-FID ($\downarrow$) \\ \hline
% 120 frames  & 20.05 & 8.511 & 0.557 \\
% 250 frames  & 21.16 & 8.654 & 0.605 \\
% 750 frames  & 21.32 & 8.655 & 0.585 \\ 
% 1500 frames & 20.60 & 8.504 & 0.522 \\\hline
% \end{tabular}
% \end{table}

% \begin{table}[]
% \caption{Ablation comparison on varying-time generation.}
% \label{ablation_longtime}
% \centering
% \begin{tabular}{ccccc}
% Length & FID ($\downarrow$) & Sync-C ($\uparrow$) & E-FID ($\downarrow$) \\ \hline
% 120 frames  & 20.05 & 8.511 & 0.557 \\
% 250 frames  & 21.16 & 8.654 & 0.605 \\
% 750 frames  & 21.32 & 8.655 & 0.585 \\ 
% 1500 frames & 20.60 & 8.504 & 0.522 \\
% \end{tabular}
% \end{table}

\begin{table}[]
\centering
\tabcolsep=0.05\linewidth
\begin{tabular}{@{}cccc@{}}
Length                   & FID ($\downarrow$) & Sync-C ($\uparrow$) & E-FID ($\downarrow$) \\ \midrule
120frames        &   16.78     &     8.25      &   0.60    \\ 
250frames &   16.96     &      8.21     &   0.67    \\ 
750frames &   16.93     &    8.46       &  0.62     \\ 
1500frames &  16.97    &    8.40       &     0.61
\end{tabular}
\caption{Comparison on long-time generation on HDTF dataset.}
\label{ablation_longtime}
\end{table}

The results indicate that as inference duration increases, the FID, SyncNet, and E-FID metrics exhibit relatively minor fluctuations without degradation trend over time, confirming the stability of EmotiveTalk in long-time inference scenarios.

\begin{figure}[t]
    \centering
    \includegraphics[width=1.0\linewidth]{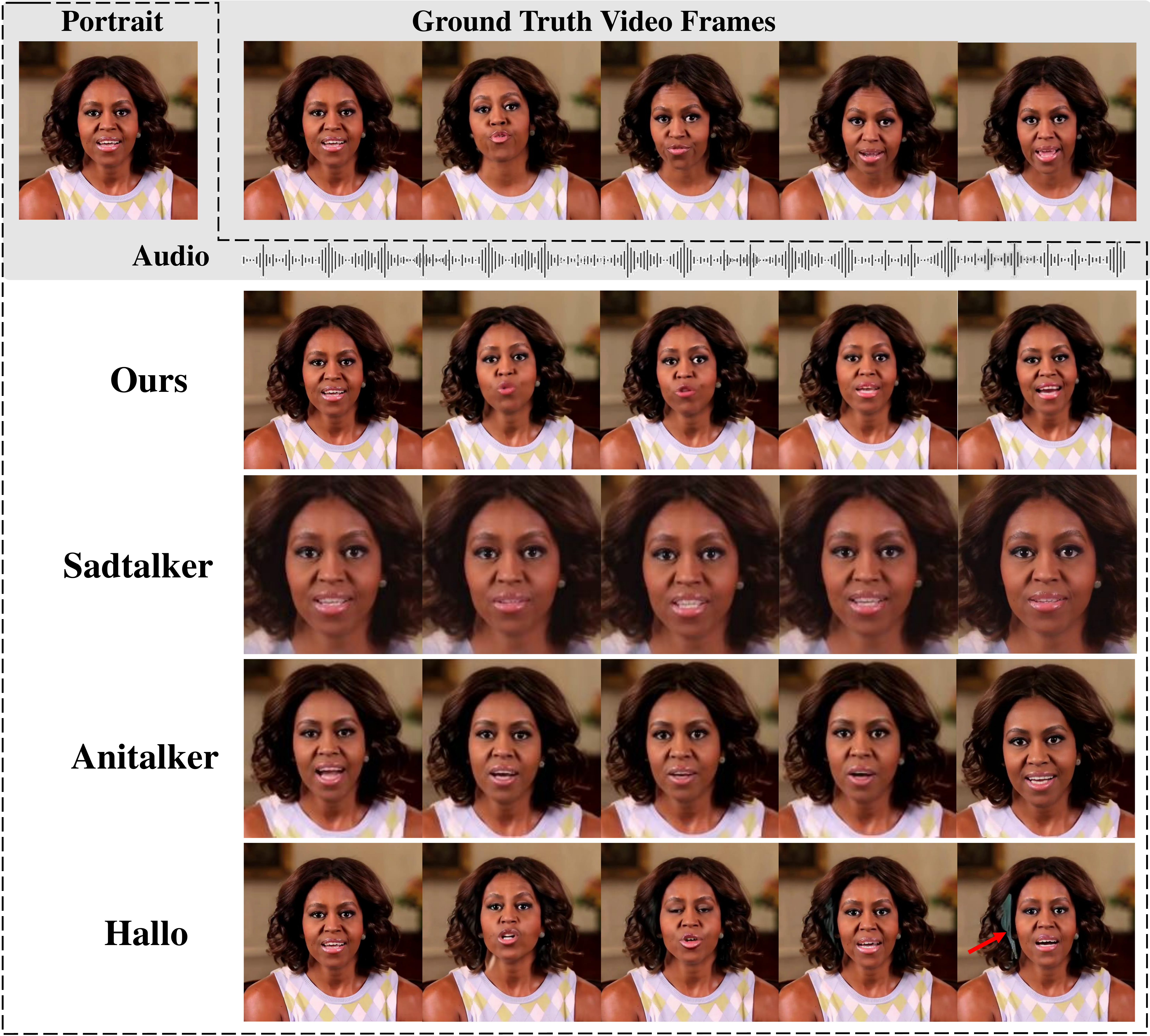}
    \caption{Case study on audio-only driven approaches.}
    \label{fig:audio_only_compare}
\end{figure}

\subsection{Case Study}
\noindent\textbf{Comparison on Audio-only Driven.} Fig.~\ref{fig:audio_only_compare} shows the qualitative results on audio-driven approaches, where driven sources contain only portrait $\mathit{\boldsymbol{x}_{\text{dri}}}$ and driven audio $\mathit{\boldsymbol{A}}$. Results show that Anitalker and Sadtalker struggle with generating video faithful to the reference image $\mathit{\boldsymbol{x}_{\text{ref}}}$ deal to the cropping and warping operation and also fall short in lip synchronization. Hallo demonstrates the ability to preserve speaker identity, but encounters instability issues in video generation, resulting in the unintended appearance of artifacts. Our method surpasses previous approaches in achieving lip synchronization, identity maintenance, and generation stability, resulting in the best overall performance.

\begin{figure}[t]
    \centering
    \includegraphics[width=1.0\linewidth]{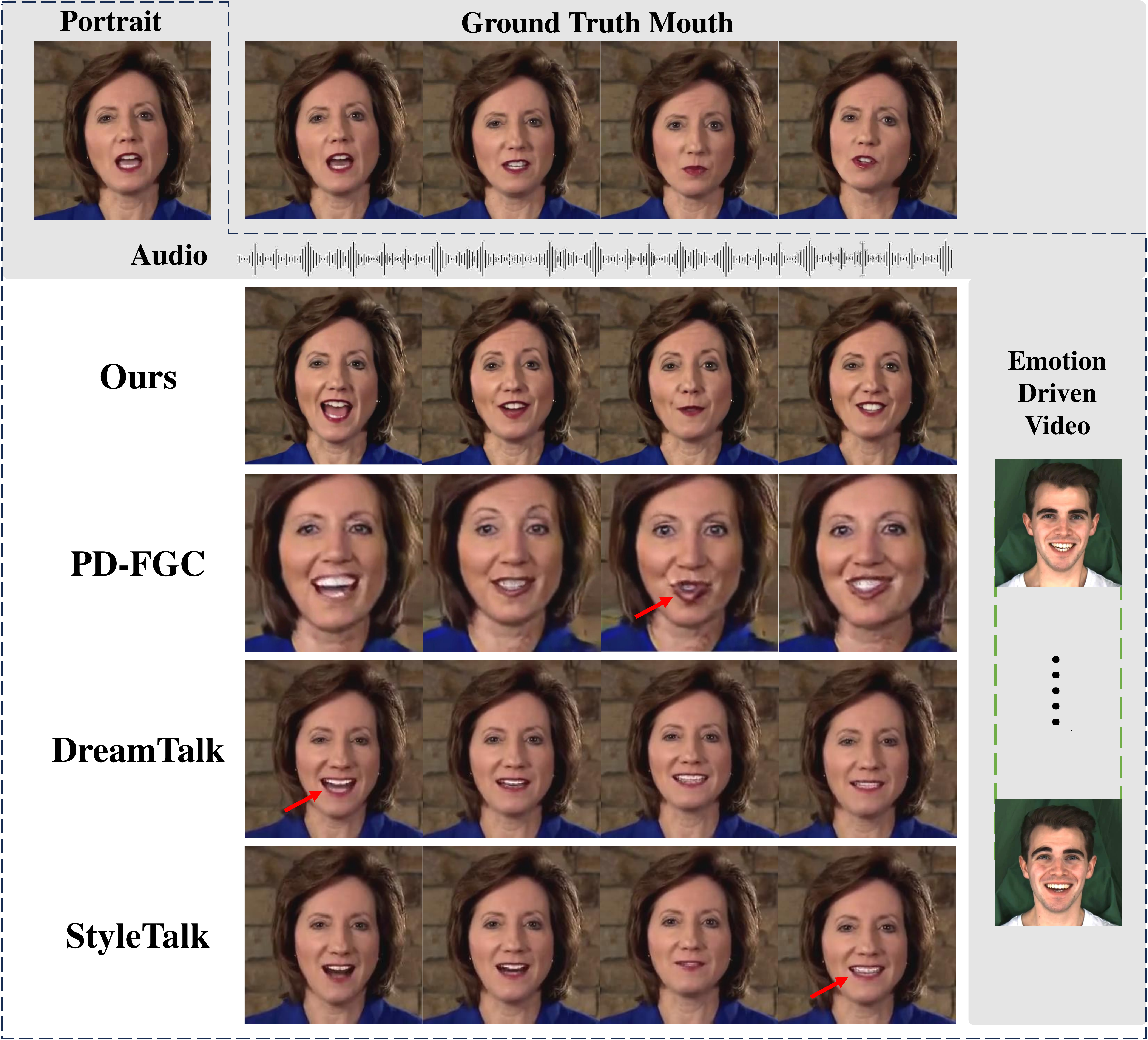}
    \caption{Case study on emotion control approaches.}
    \label{fig:video_driven_compare}
\end{figure}

\noindent\textbf{Comparison on Emotion Control.} To evaluate the performance of emotion control, we use a portrait paired with a happy video from another person and employ various methods to transfer the emotion. Fig.~\ref{fig:video_driven_compare} shows the results, which indicate that StyleTalk and DreamTalk struggle in lip synchronization due to the coupling of lip and expression. PD-FGC faces the challenge of lip shape deformation. Our method achieves the most neutral and expressive emotion control results
also ensures lip sync, highlighting the effectiveness of our decoupling approach and model design.

\begin{table}[t] 
\centering
\resizebox{0.47\textwidth}{!}{
\begin{tabular}{@{}ccccc@{}}
Methods                   & Lip-Sync ($\uparrow$) & Exp-Q ($\uparrow$) & Realness ($\uparrow$) & V-Q ($\uparrow$) \\ \midrule
SadTalker~\cite{sadtalker}&   3.03     &   3.03    &    3.01       &  3.29    \\
Anitalker~\cite{anitalker}&   2.82    &    3.04   &    2.87       &   3.24    \\
Aniportrait~\cite{aniportrait}&  1.65   &   1.79     &     1.65      &   2.26   \\
Hallo~\cite{hallo}&   3.73        &    3.36    &     3.28      &   3.49    \\
StyleTalk~\cite{styletalk}&  2.50   &   2.88     &    2.78       &    3.02   \\
DreamTalk~\cite{dreamtalk}&  3.69   &  3.45      &     3.40      &   3.38    \\
\rowcolor{green!8}
Ours&  \textbf{4.15}   &   \textbf{3.96}     &     \textbf{3.98}      &    \textbf{4.03}   \\
Ground Truth&  4.51   &   4.49     &   4.44        &      4.40
\end{tabular}}
\caption{User Study Results.}
\label{tab:user_study}
\end{table}

\subsection{User Study}
\label{sec:userstudy}
We generated 10 test samples covering various emotion states and used 7 different models to generate with the ground-truth samples included. We conducted a user study of 26 participants, for each method, the participant is required to score 10 videos sampled from the test samples and is asked to give a rating (from 1 to 5, 5 is the best) on four aspects: (1) the lip sync quality (Lip-Sync), (2) the quality of expressions (Exp-Q), (3) the realness of results (Realness), (4) the quality of generated video (V-Q). The results are shown in Tab.~\ref{tab:user_study}, our method outperforms existing approaches across all aspects, particularly in expression quality and lip sync, highlighting its superior capabilities.

\section{Conclusion}
In this work, we propose EmotiveTalk, a novel method that aims at enhancing the emotional expressiveness and better controllability of talking head video generation. We propose a novel approach to decouple audio embedding by leveraging facial motion information, enabling the generation of decoupled representations that correspond directly to lip motions and facial expressions. Additionally, we introduce a well-designed video diffusion framework that drives these representations to generate expressive and extended talking head videos. Due to the effectiveness of our decoupling, we further enhance the controllability of emotion generation by incorporating additional emotion information from multiple optional sources to customize the generated emotions. Extensive experiments demonstrate the superiority of EmotiveTalk.

%% file: sec/X_suppl.tex
\clearpage
\setcounter{page}{1}
\maketitlesupplementary

\begin{figure}[t]
  \centering
  \includegraphics[width=1.0\linewidth]{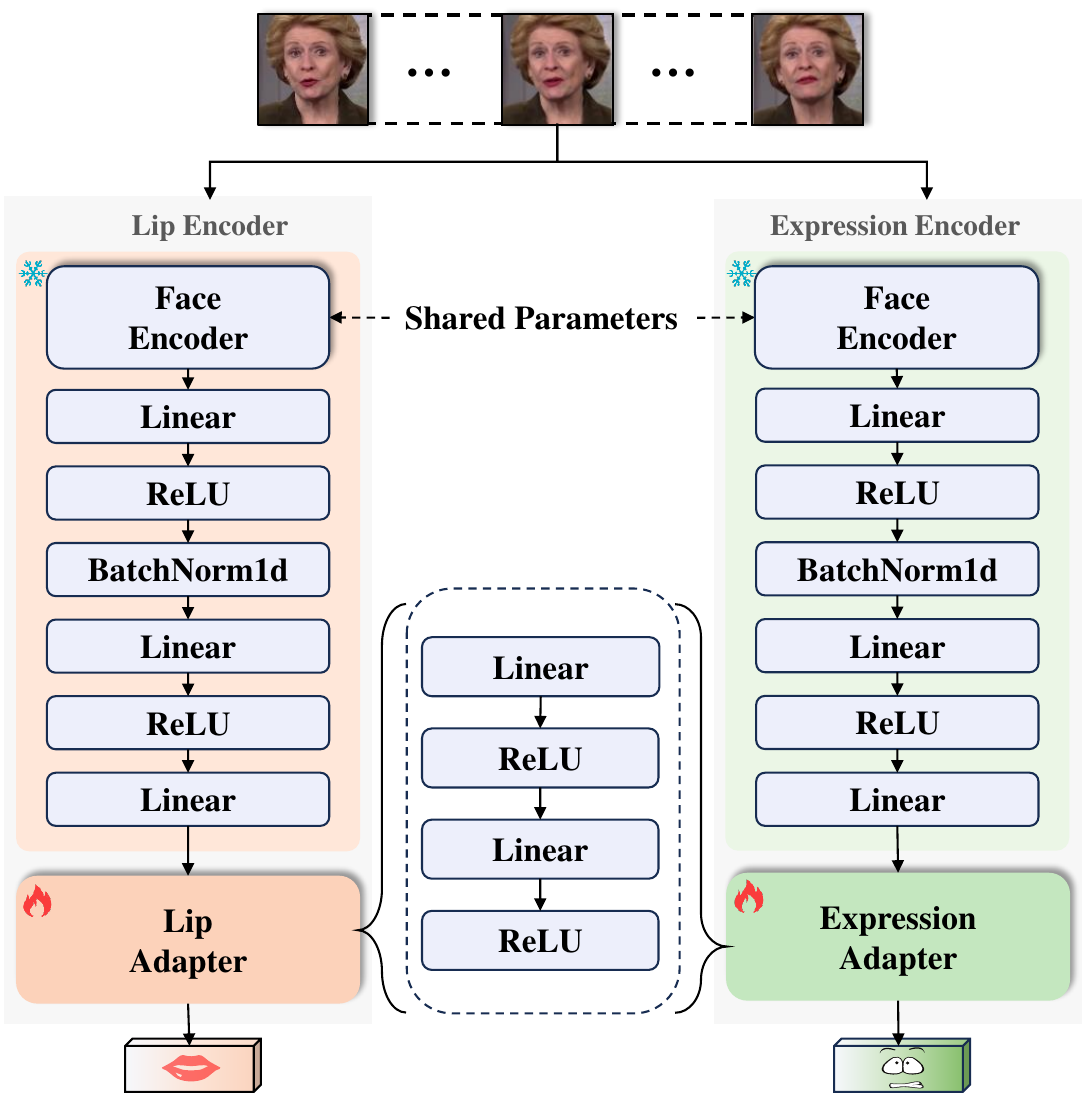}
  \caption{The structure of lip and expression encoder in EmotiveTalk.}
  \label{fig:pdfgcencoder}
\end{figure}

\begin{figure}[]
  \centering
  \includegraphics[width=1.0\linewidth]{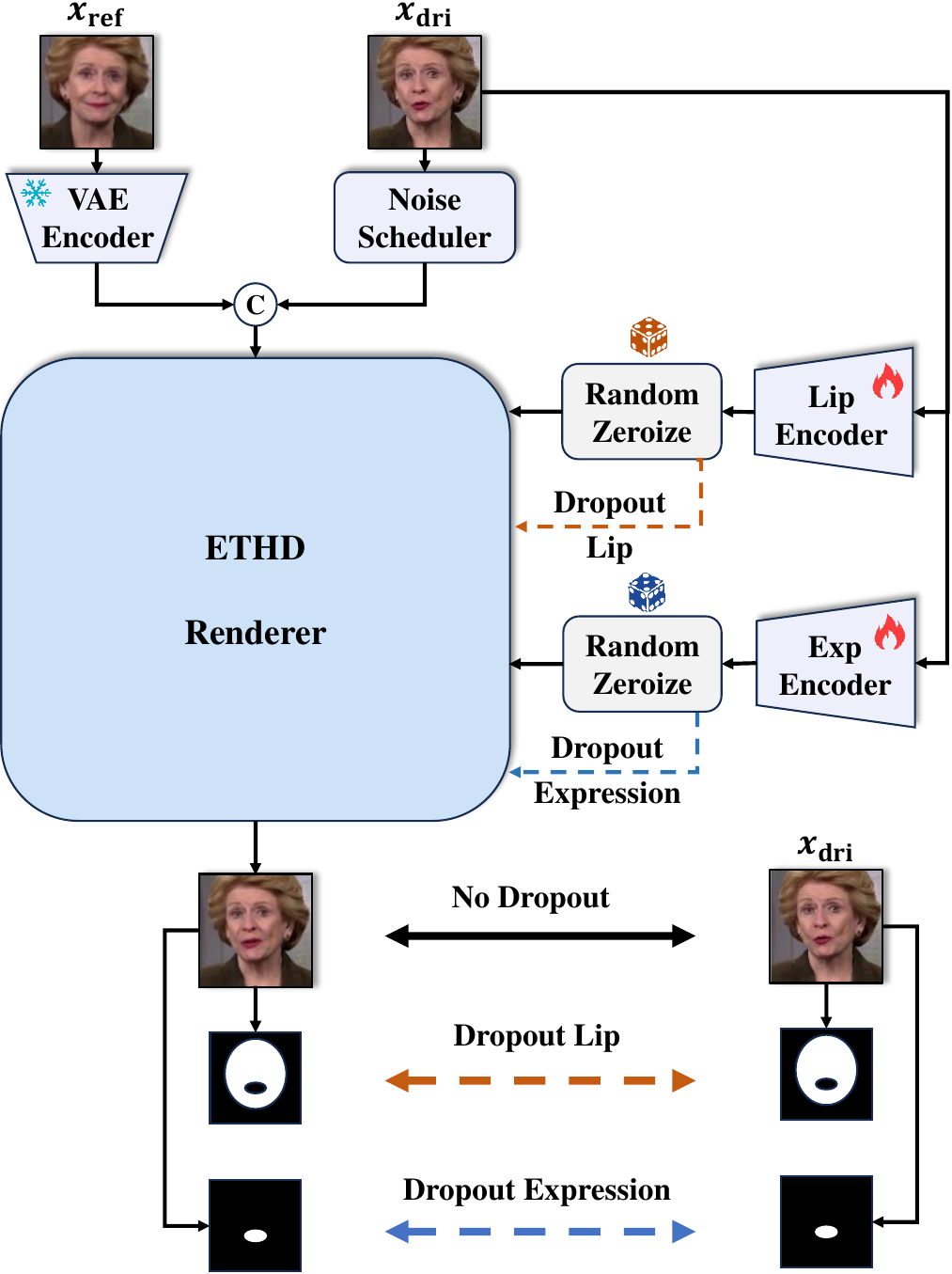}
  \caption{The self-driven dropout image reconstruction training process for lip and expression encoders pre-training.}
  \label{fig:droupout}
\end{figure}

\section{Implementation Details of EmotiveTalk}

\subsection{Vision Encoders Pre-training}
\label{sec:pre-training}
In talking head videos, the movements of different facial regions are often coupled, which increases the difficulty of independently controlling facial expressions and lip motions. Extracting decoupled representations related to expressions and lips is crucial for generating controllable expressions in talking head models. To obtain better performance and controllability in emotional talking head generation, we design lip and expression encoders to encode the lip motions and facial expressions independently to generate lip motion latent embedding $\mathit{\boldsymbol{\Vec{l}}}_\text{v}$ and expression latent embedding $\mathit{\boldsymbol{\Vec{e}}}_\text{v}$ from self-driven video. 

\noindent\textbf{Architecture of Lip and Expression Encoders.} Our lip and expression encoders are based on PDFGC~\cite{pdfgc}, which demonstrates remarkable facial motion decoupling results. We add trainable lip and expression adapters before the pre-trained PDFGC encoder to achieve better decoupling results based on the following tasks. Our network structure of lip and expression encoders are illustrated in~\cref{fig:pdfgcencoder}.

\noindent\textbf{Self-driven Dropout Image Reconstruction}. We have pre-trained our lip and expression encoders on a self-driven task. We employ the ETHD backend described in ~\cref{sec:ETHD} as the renderer to perform self-driven image reconstruction. The inputs consist of a reference image and another driving image from the same speaker. The driving image is processed by the lip and expression encoder to obtain latent embeddings of the lip and expression, which serve as conditional inputs to the renderer.

To more effectively combine the driving representations for controlling movements in different facial regions and to alleviate the coupling between lips and expressions, we adopt condition dropout training. As shown in Fig.~\ref{fig:droupout}, during training, we adopt a conditional dropout strategy with three configurations:
\begin{itemize}
\item Dropping the lip latent embedding $\mathit{\boldsymbol{\Vec{l}}}_\text{v}$ ;
\item Dropping the expression latent embedding $\mathit{\boldsymbol{\Vec{e}}}_\text{v}$; 
\item Utilizing both latent embeddings simultaneously.
\end{itemize}
When dropping the expression latent embedding, we apply the facial mask described in ~\cref{sec:EDI} to both the ground-truth and the generated frame latents, ensuring that parameter updates focus exclusively on the reconstruction of facial expressions. Similarly, when dropping the lip latent embedding, we apply the lip mask from ~\cref{sec:EDI} to both ground-truth and the generated frame latents and focus only on lips reconstruction. When both latent embeddings are used simultaneously, we train the model to reconstruct the entire image.

\begin{figure}[t]
  \centering
  \includegraphics[width=1.0\linewidth]{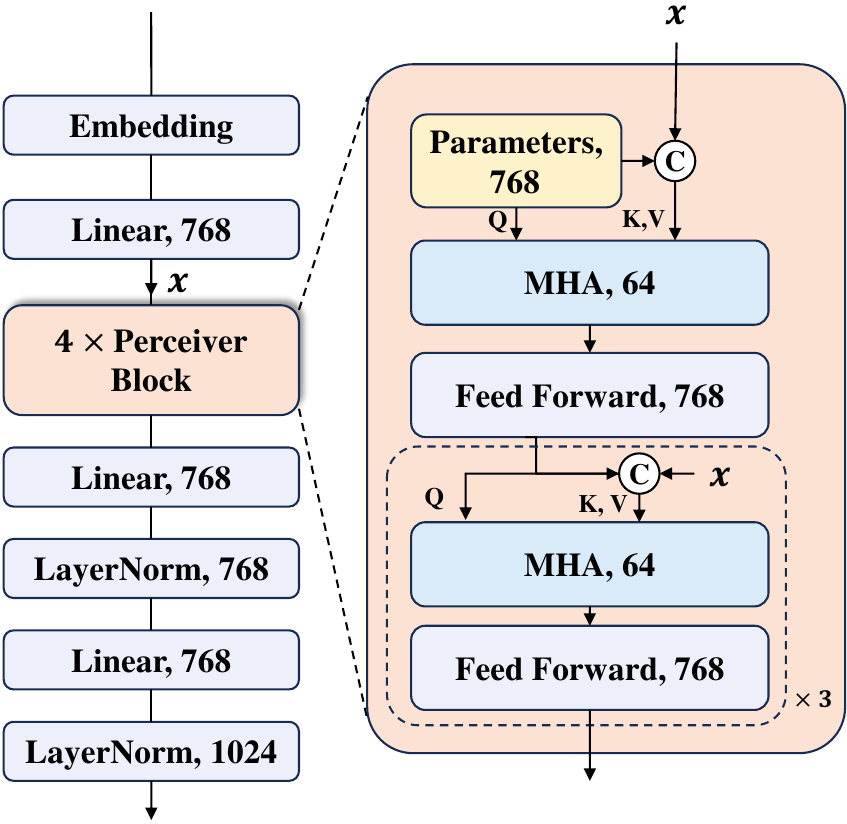}
  \caption{The structure of the audio-lip projector of V-AID in EmotiveTalk.}
  \label{fig:perceiver}
\end{figure}

\subsection{Details of V-AID}
\noindent\textbf{Architecture of Audio-lip Projector.} Our audio-lip projector demonstrated in ~\cref{sec:V-AID} leverages a Perceiver Transformer~\cite{perceiver} architecture, illustrated in Fig.~\ref{fig:perceiver}. The input to the audio-lip projector consists of audio embedding window $\mathit{\boldsymbol{A}}_{\text{w}} \in \mathbb{R}^{l \times w \times c}$ encoded by a pre-trained Wav2Vec~\cite{wav2vec} encoder, where $l$ denotes the length of audio and $w$ denotes the window size. The audio embedding is first passed through an embedding layer and a linear layer for feature projection. Subsequently, the processed embedding $x$ is fed into the following four Perceiver Transformer blocks. In the first block, the query input $y$ is a learnable vector, while the keys and values are derived by concatenating $y$ with the original input $x$. After a matrix transformation, multi-head attention is applied along the window dimension to capture internal relationships within the speech feature window. The output of this process is further refined using a feed-forward module. As for the following blocks, the query input is the output of the preceding block, while the keys and values are constructed by concatenating the block's output with the original input $x$. This design iteratively enhances the feature representations by leveraging both the temporal dependencies and the shared information between the block outputs and the original input. Leveraging the vision-guided training strategy described in ~\cref{sec:V-AID}, the model effectively learns to map audio representation space to the corresponding lip motion representation space. 

\subsection{Details of MEC}

\begin{figure}[t]
  \centering
  \includegraphics[width=1.0\linewidth]{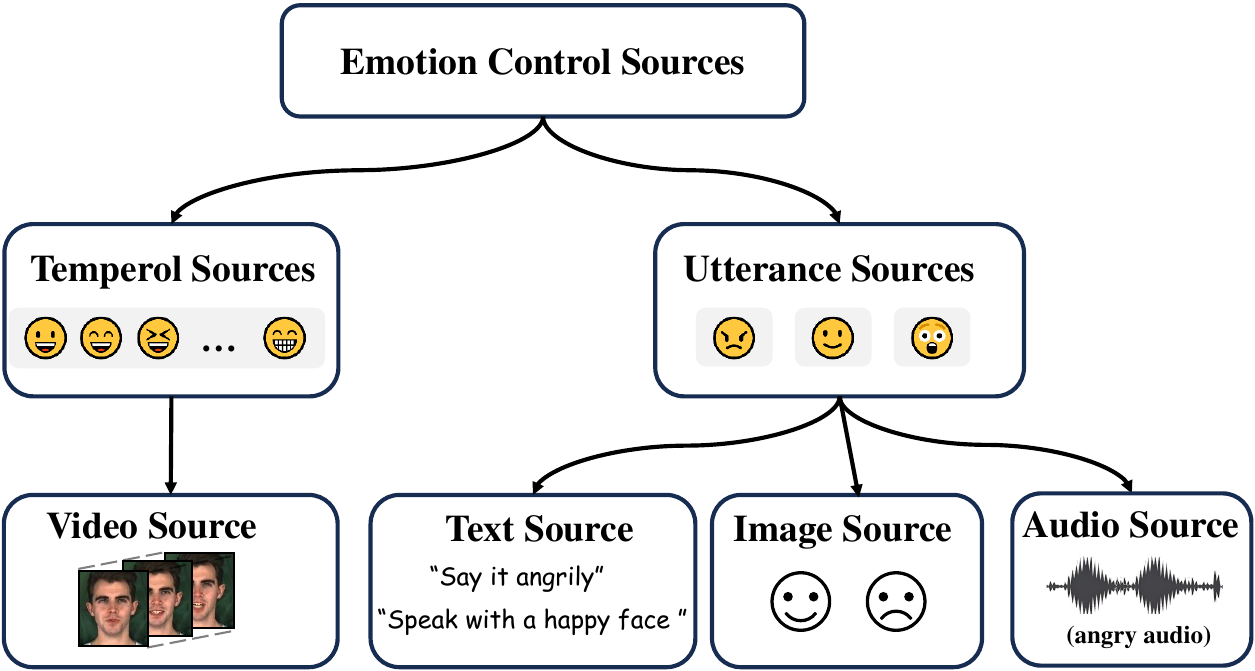}
  \caption{The categories of emotion control sources.}
  \label{fig:mec}
\end{figure}

\begin{figure}[t]
  \centering
  \includegraphics[width=1.0\linewidth]{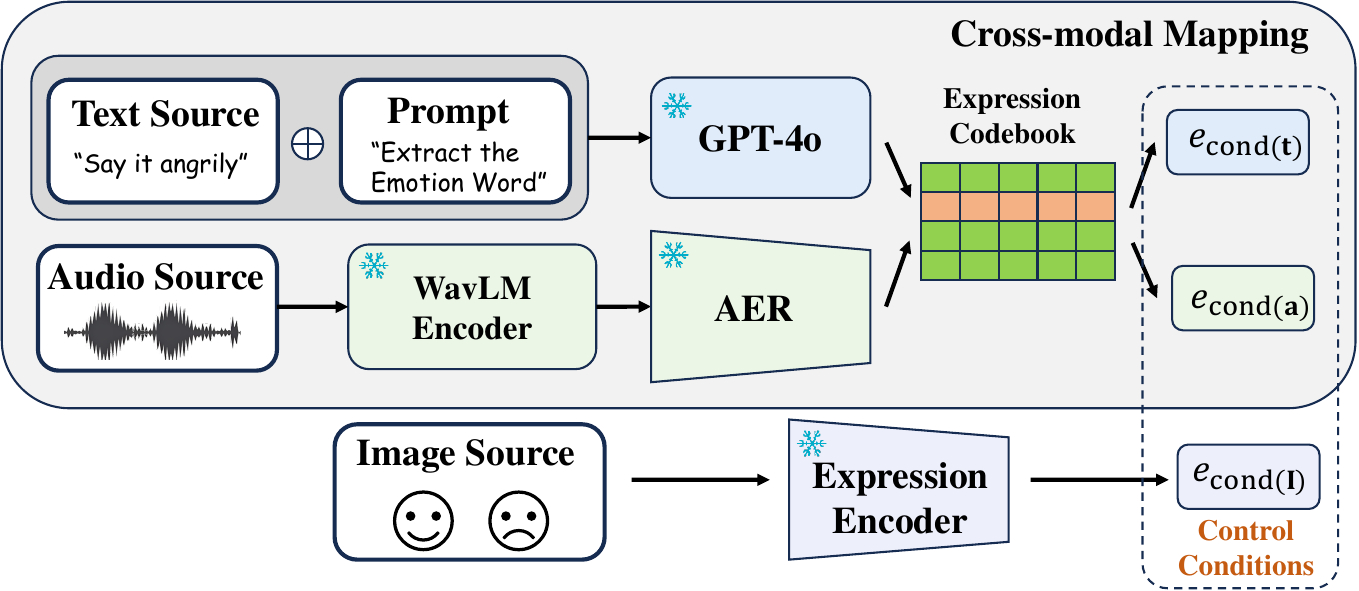}
  \caption{The designed mapping module of mapping different utterance sources to vision expression space to serve as emotion control conditions.}
  \label{fig:crossmodal}
\end{figure}

To facilitate the use of multiple optional emotion-driven control sources for customizing speaking emotions, we propose a Multi-Source Emotion Control (MEC) pipeline tailored to different emotion control sources. This section provides a detailed elaboration of the concepts introduced in ~\cref{sec:MEC}. Overall, the pipeline incorporates various types of emotion control inputs, as illustrated in Fig.~\ref{fig:mec}, for temporal sources like driven video, we utilize the pre-trained expression encoder to obtain expression latent embedding from the emotion video source as the expression-driven representation to control the expression generation, as illustrated in ~\cref{sec:MEC}. As for the utterance sources, we map the emotional information to the emotion condition $\mathit{\boldsymbol{e}_\text{cond}}$, then we utilize the Di-CTE model in ~\cref{sec:V-AID} to expanding the utterance condition to temporal driven condition. In this section, we provide a detailed explanation of how emotion conditions are derived from various utterance control sources, as illustrated in~\cref{fig:crossmodal}.

\noindent\textbf{Image Source.} We utilize the pre-trained expression encoder to obtain expression latent embedding from the driven image as the emotion control condition, as illustrated in ~\cref{sec:MEC}.

\noindent\textbf{Text Source.} We utilize GPT-4o to retrieve emotion keywords from specified emotion prompts. To facilitate this process, we define a set of eight commonly observed emotion keywords: happy, angry, sad, surprised, fear, disgusted, worried, and neutral. Based on these keywords, we construct an emotion-to-expression codebook, which stores latent representations of facial expressions corresponding to each emotion. These latent representations are extracted using the pre-trained expression encoder (~\cref{sec:pre-training}) applied to facial images reflecting various emotional states. To enhance the extraction process, we prepend a retrieval prompt to the emotion prompt, which is ``Extract the keyword representing emotion from the following sentence and return only the keyword. The keywords should be one of happy, angry, sad, surprised, fearful, disgusted, worried, or neutral." 

\noindent\textbf{Audio Source.} To enable the retrieval of emotion-driven keywords from external emotional audio input, we have designed an audio-emotion retrieval module. This module employs a frozen audio encoder and a trainable speech emotion recognition module. In our implementation, we utilize WavLM as the audio encoder due to its superior performance in speech emotion recognition tasks. The architecture of our speech emotion recognition module leverages the audio branch of the state-of-the-art multimodal emotion recognition model. Similarly, we use an emotion-to-expression codebook to map the retrieved emotional keywords to latent embedding representations in the expression space. These representations serve as emotional control conditions for the subsequent Di-CTE module, enabling the generation of sequential emotional control conditions.

\subsection{Training and Inference Details}
\label{sec:details of training and inference}
This section provides a detailed explanation of the training and inference processes for EmotiveTalk, serving as a supplementary discussion to~\cref{sec:training and inference}. 

\noindent\textbf{Training Details.} EmotiveTalk is trained using eight NVIDIA A100 GPUs. Compared to the current mainstream talking head video generation models based on the Stable Diffusion~\cite{latentdiffusion} framework, our model's initialization weights are partially inherited from the pre-trained Stable Video Diffusion model~\cite{stablevideodiffusion}. Moreover, our model achieves a balance between performance and efficiency, supporting the training of long-duration, high-resolution videos. In our project, we train the audio-lip projector and Di-CTE module in V-AID in ~\cref{sec:V-AID} with a learning rate of $1\times10^{-4}$ and a batch size of 16 for each iteration by Adam optimizer and training length of each expression-related latent Di-CTE module is set to 220, and randomly cat the ground-truth expression latent for a ratio of 0.8. We train the ETHD backbone network under the following three configurations:

\begin{itemize}
\item 512-resolution: Training images and videos at a resolution of $512\times512$, with 120 frames per training iteration; 
\item 1024-resolution: Training images and videos at a resolution of $1024\times1024$, with 32 frames per training iteration.
\end{itemize}

The training length of 1024-resolution is much shorter than 512-resolution due to the training cost constraint of high resolution. The hyperparameters for all configurations are kept consistent. We use the Adam8bit optimizer with a learning rate of $1\times10^{-5}$ and a batch size of 1 for each iteration. And the possibility of choosing ground-truth vision expression latents $\mathit{\boldsymbol{\Vec{e}}}_{\text{v}}$ when training is set to 0.6.

\noindent\textbf{Inference Details.} During inference, we first utilize the V-AID module to generate lip-related latent embeddings $\mathit{\boldsymbol{l}_\text{a}}$ and expression-related latent embeddings $\mathit{\boldsymbol{e}_\text{a}}$ based on the driving speech signal and emotional control source. For long-time expression-related latent embedding generation, we guide the inference process by appending the last 20 frames of the generated expression-related latent embeddings to the sampled noise as initialization.

Subsequently, the lip-related $\mathit{\boldsymbol{l}_\text{a}}$ and expression-driven latent embeddings $\mathit{\boldsymbol{e}_\text{dri}}$ obtained in the previous step are used as conditional inputs for the denoising process of the ETHD backbone network. $\mathit{\boldsymbol{e}_\text{dri}}$ is choose from $\mathit{\boldsymbol{e}_\text{a}}$ and $\mathit{\boldsymbol{e}_\text{v}}$ based on the category of the emotion source, $\mathit{\boldsymbol{e}_\text{v}}$ for temporal sources and $\mathit{\boldsymbol{e}_\text{a}}$ for utterance sources. We adopt the DDIM inference strategy with 25 denoising steps. The inference configurations for different resolutions are as follows:

\begin{itemize}
\item 512-resolution: Window size of 120 frames with an overlap of 24 frames; 
\item 1024-resolution: Window size of 32 frames with an overlap of 12 frames.
\end{itemize}

\section{Detailed Evaluations of EmotiveTalk}
\label{sec:moreexperiments}

This section serves as an extended discussion of~\cref{sec:experiments} in the main paper, providing a more comprehensive and detailed analysis and comparison of EmotiveTalk.

\subsection{Evaluation Details of Our Results in the Main Paper}
This section serves as a detailed explanation of the evaluation settings of our experiments in~\cref{sec:experiments} of the main paper. 
\noindent\textbf{Evaluation Settings Details.} We utiliz the publicly available HDTF~\cite{hdtf} and MEAD~\cite{mead} datasets for training and evaluation. The same data split is applied to both datasets: 90\% of the data is allocated to the training set for model training, while the remaining 10\% is reserved as the evaluation set for evaluation. To ensure robust evaluation, we strictly maintain no overlap between the training set and the evaluation set, and the evaluation set data is entirely unseen during the training process. Specifically, for the MEAD dataset, due to limitations in training length, we filter the training set by excluding sequences shorter than 120 frames (equivalent to 4.8 seconds). For evaluation, MEAD sequences shorter than 120 frames are zero-padded to reach this duration of 120 frames. When generating videos from the processed test data, frames beyond the original sequence length are removed using the ffmpeg tool, ensuring the generated videos match the length of the original ground-truth video. We utilize these above-mentioned principles for all our evaluations in the main paper~\cref{sec:training and inference} and following experiments. 

\noindent\textbf{Comparison Settings Details.} To ensure a fair comparison, we conduct an evaluation with audio of the same length when comparing our model with other state-of-the-art models in~\cref{audionly_results}. On the HDTF dataset, we use audio clips of 5.76 seconds (approximately 144 frames in the ground-truth video) to drive the portrait. Since different models generate slightly varying numbers of frames for the same audio length, we standardize the evaluation by reporting the FVD metric for the first 128 frames ($\text{FVD}_{\text{128}}$), along with the average FID, and Sync-C Sync-D metrics across all generated frames. As for the E-FID metric, we follow the approach used in EMO~\cite{emo}, extracting 3D reconstructed expression coefficients from all frames of both the generated and ground truth videos. The E-FID is then computed as the FID between the expression coefficients of the generated and ground truth videos. And on the MEAD dataset, due to the shorter video length, we standardize the evaluation by generating 3.04 seconds (approximately 76 frames in the ground-truth video) and testing the FVD metic for the first 72 frames ($\text{FVD}_{\text{72}}$). The settings of other metrics are the same as those of the abovementioned HDTF testing. This methodology ensures the comprehensiveness and fairness of the evaluation, providing an objective comparison across all the models.

\noindent\textbf{Users Study Details.} As described in~\cref{sec:userstudy}, our user study involved 26 participants, including 10 professional video evaluation engineers and 16 graduate students with experience in audio and video information processing. We select 10 video clips from the partitioned HDTF evaluation set and extract their audio to generate video clips using the models outlined in~\cref{sec:userstudy}. For audio-video driven models that require video input, such as DreamTalk~\cite{dreamtalk} and StyleTalk~\cite{styletalk}, the original videos are also provided as inputs to these models. In contrast, our model is operated solely in an audio-driven manner without incorporating additional emotion control conditions, ensuring fairness in input information. We employed several subjective evaluation criteria and defined detailed quantitative metrics for each score level in the subjective standards. These requirements are thoroughly documented in an evaluation guide provided to the participants. Our designs ensured the validity and fairness of the user study.

\begin{figure}[t]
  \centering
  \includegraphics[width=1.0\linewidth]{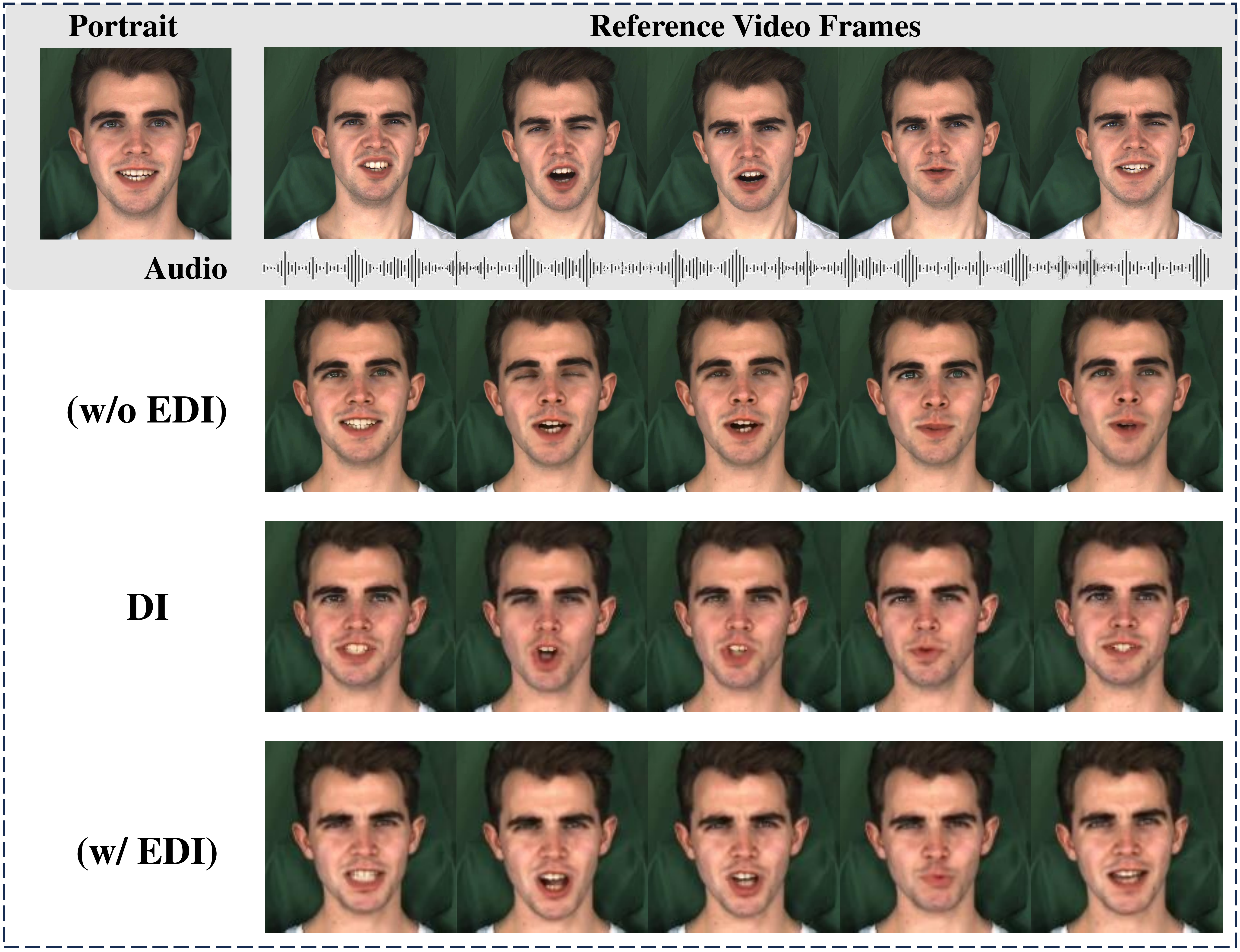}
  \caption{Ablation studies on EDI block.}
  \label{fig:edi}
\end{figure}

\begin{figure}[t]
  \centering
  \includegraphics[width=1.0\linewidth]{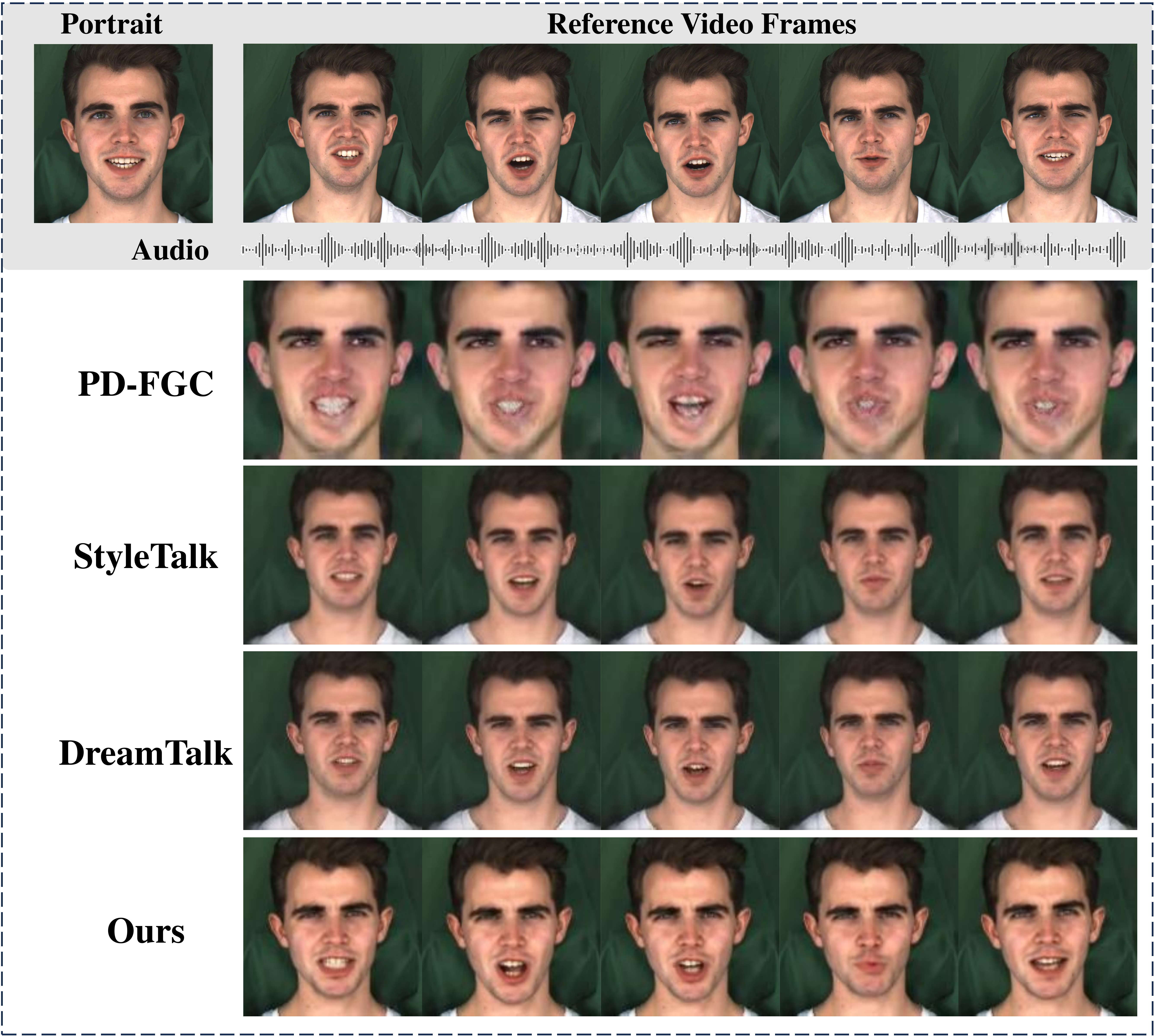}
  \caption{Comparison of emotion transferring with other methods.}
  \label{fig:emoctrl}
\end{figure}

\begin{table*}[t]
\renewcommand{\arraystretch}{1.2}
\centering
\tabcolsep=0.02\linewidth
\renewcommand\arraystretch{1.6}
\begin{tabular}{c|cccccc}
\multirow{2}{*}{Methods} & \multicolumn{6}{c}{HDTF / MEAD}                                                            \\
                         & Driven & FID ($\downarrow$)         & FVD ($\downarrow$)          & Sync-C ($\uparrow$)   & Sync-D ($\downarrow$)     & E-FID ($\downarrow$)    \\ \hline
Ours (512) & A & 16.64 / 53.21          & 140.96 / \textbf{207.67} & \textbf{8.24} / \textbf{6.82} & \textbf{7.09} / \textbf{7.43} & \textbf{0.54} / 0.57 \\ 
Ours (1024) & A & \textbf{12.75} / \textbf{37.97}          & \textbf{134.64} / 303.55 & 7.04 / 4.60 & \ 8.09 / 9.58 & 0.62 / \textbf{0.54} \\\midrule\midrule
Ours (512) & A+V & 16.09 / 50.84 & {\textbf{120.70} / \textbf{153.71}}    & \textbf{8.41} / \textbf{6.79}       & \textbf{7.11} / \textbf{7.58}       & {\textbf{0.34} / \textbf{0.40}} \\
Ours (1024) & A+V & \textbf{12.68} / \textbf{38.33} & {147.40 / 275.04}    & 7.85 / 5.00      & 7.82 / 9.55      & {0.42 / 0.43} \\\midrule\midrule
Ground Truth             & A+V   & -         & -          & 8.63 / 7.30       & 6.75 / 8.31         & -      
\end{tabular}
\caption{Comparisons of our models with different resolutions on HDTF and MEAD. ``A" denotes audio-driven and ``A+V" denotes audio-video driven. ``$\uparrow$'' indicates better performance with higher values, while ``$\downarrow$'' indicates better performance with lower values.}
\label{different_resolution}
\end{table*}

\subsection{Ablation studies on EDI}
The design of the EDI module aims to achieve the automatic removal of the emotion information from the reference image while injecting target emotion information into the hidden states during both training and inference. This enables the transformation of facial expressions from the reference image to the target expression. To evaluate the effectiveness of the EDI module, we conduct the following ablation experiment. The experiments are performed on the test set of MEAD, as described in~\cref{sec:experiments}, due to the extensive diversity of speaker emotions in the MEAD dataset, with three different configurations:

\begin{itemize}
\item Without the EDI module (w/o EDI); 
\item Direct injection of target emotional information (DI);
\item Using the EDI module for emotional information injection (w/ EDI).
\end{itemize}

The results are presented in~\cref{fig:edi}. The results demonstrate that without utilizing the EDI module (w/o EDI), it is challenging to achieve a transition from happy to angry in the emotion of generated video, even when provided with video information depicting anger. The direct injection approach (DI) fails to effectively eliminate the influence of the happy emotion from the reference image, resulting in the residual coupling of happiness in the generated images and weakening the expression of anger. In contrast, the conditional injection method using the EDI module effectively removes the residual coupling of happiness, enabling a more expressive and accurate transformation from the happy reference image to the generated video portraying anger.

Furthermore, to evaluate the performance of emotional state transfer, we compare our proposed method with other state-of-the-art emotion control methods to compare the results. Fig.~\ref{fig:emoctrl} shows the qualitative results on emotion control generation by emotion reference video, where ``ours'' denotes our model with EDI. Results show that StyleTalk and DreamTalk fail to preserve speaker identity. They inadvertently reveal extraneous positional information about the reference video, as the spatial positioning and head size in the generated video are aligned with the emotion reference video rather than maintaining consistency with the reference image. PD-FGC faces challenges in low-definition issues in the lip region. Compared to all the methods, our method with EDI achieves the most expressive emotion control results while preserving speaker identity, resulting in the best performance among all the methods.

\subsection{Comparison on Different Resolutions} 

To quantitatively compare the impact of different resolution training strategies on EmotiveTalk, we have trained two models at resolutions of 512 and 1024 using the configurations described in~\cref{sec:details of training and inference}. The same quantitative evaluation metrics as outlined in the main paper~\cref{sec:experiments} are used to validate performance on the evaluation set of HDTF and MEAD. The results are shown in Tab.~\ref{different_resolution}.

The results demonstrate that models trained at different resolutions exhibit distinct strengths across various metrics on both datasets. The 1024-resolution model significantly outperforms the 512-resolution model in terms of the FID metric, highlighting the superior ability of high-resolution training to better preserve image details. Conversely, the 512-resolution model achieves notably better performance on lip-sync metrics, Sync-C and Sync-D in Tab.~\ref{different_resolution}, underscoring the positive impact of long-time training on generating lip movements that are more consistent with the audio.

\begin{table}[]
\renewcommand{\arraystretch}{1.4}
\centering
\tabcolsep=0.046\linewidth
\begin{tabular}{@{}cccc@{}}
\toprule
Methods     & Params ($\downarrow$)  & TFLOPs ($\downarrow$) & Time ($\downarrow$)  \\ \midrule
Hallo~\cite{hallo}       & 2.17G & 4388.32 & 501.97 \\
Ours        & \textbf{1.58G} & \textbf{3668.16} & \textbf{435.18} \\ \bottomrule
\end{tabular}
\caption{Comparisons of network efficiency of denoising backbone with diffusion-based methods. ``Params" denotes the network parameters of the backbone, ``TFLOPs'' denotes the computation cost, and ``Time'' denotes the time cost. ``$\downarrow$'' indicates better performance with lower values.}
\label{efficiency}
\end{table}

\subsection{Comparison on Network Efficiency}

To further investigate the advantages of our model's efficient design compared to other diffusion-based talking head generation models, we conducted a comprehensive evaluation from two perspectives: model parameters and inference computational cost. We include a representative method Hallo~\cite{hallo} that is similar to our method, which also employs a 3D U-Net architecture and utilizes pre-trained models of Stable Diffusion~\cite{latentdiffusion}. For a fair comparison, we test the total parameters of the backbone network and computational flops of generating the same length of talking head video (4.8s for each model exactly) with the inference settings of each model, respectively. The comparison results are summarized in Tab.~\ref{efficiency}.

From the metrics, it is evident that our method demonstrates clear advantages over the mainstream reference-denoising UNet architecture, represented by Hallo~\cite{hallo}, in terms of parameter count, computational cost, and generation time. Furthermore, as discussed in~\cref{audionly_results} of the main paper, our approach also outperforms Hallo in terms of performance on most of the evaluated metrics. This highlights the efficiency of our network design, achieving a balance between performance and efficiency while maintaining superior performance.

\begin{figure*}[t]
  \centering
  \includegraphics[width=1.0\linewidth]{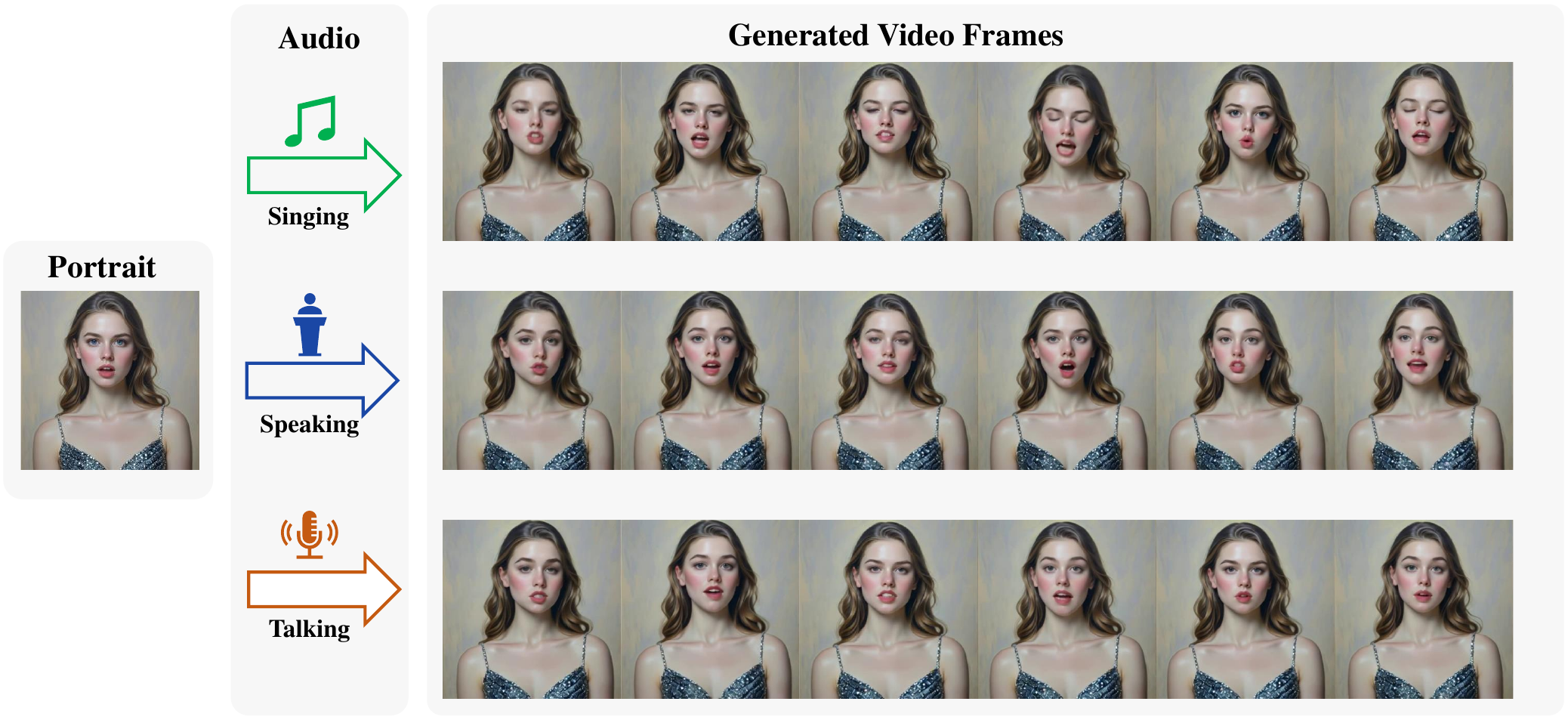}
  \caption{Video generation results of EmotiveTalk given different audios of variable styles.}
  \label{fig:multistyle}
\end{figure*}

\subsection{Additional Generation Results}

\noindent\textbf{Results on Multiple Audio Styles.} We conduct experiments to compare the driving effects of different audio styles on the same reference portrait by generating videos using varying audio inputs. Three distinct audio styles are employed:

\begin{itemize}
\item Singing: Driven by English song audio; 
\item Speaking: Driven by English speech recordings;
\item Talking: Driven by English daily talking audio.
\end{itemize} 

The results are shown in~\cref{fig:multistyle}. In the visualized results, we observe that using singing audio introduces more frequent blinking, expressive facial changes, and rhythmic head movements. Using speaking driven achieves a vivid and dynamic result, characterized by rich facial expressions, pronounced mouth movements, and noticeable head movements. On the other hand, the result driven by daily talking audio achieves realistic and natural driving effects. These visualized results demonstrate the effectiveness of our proposed method in handling diverse audio styles for driving animations.

\begin{figure*}[t]
  \centering
  \includegraphics[width=1.0\linewidth]{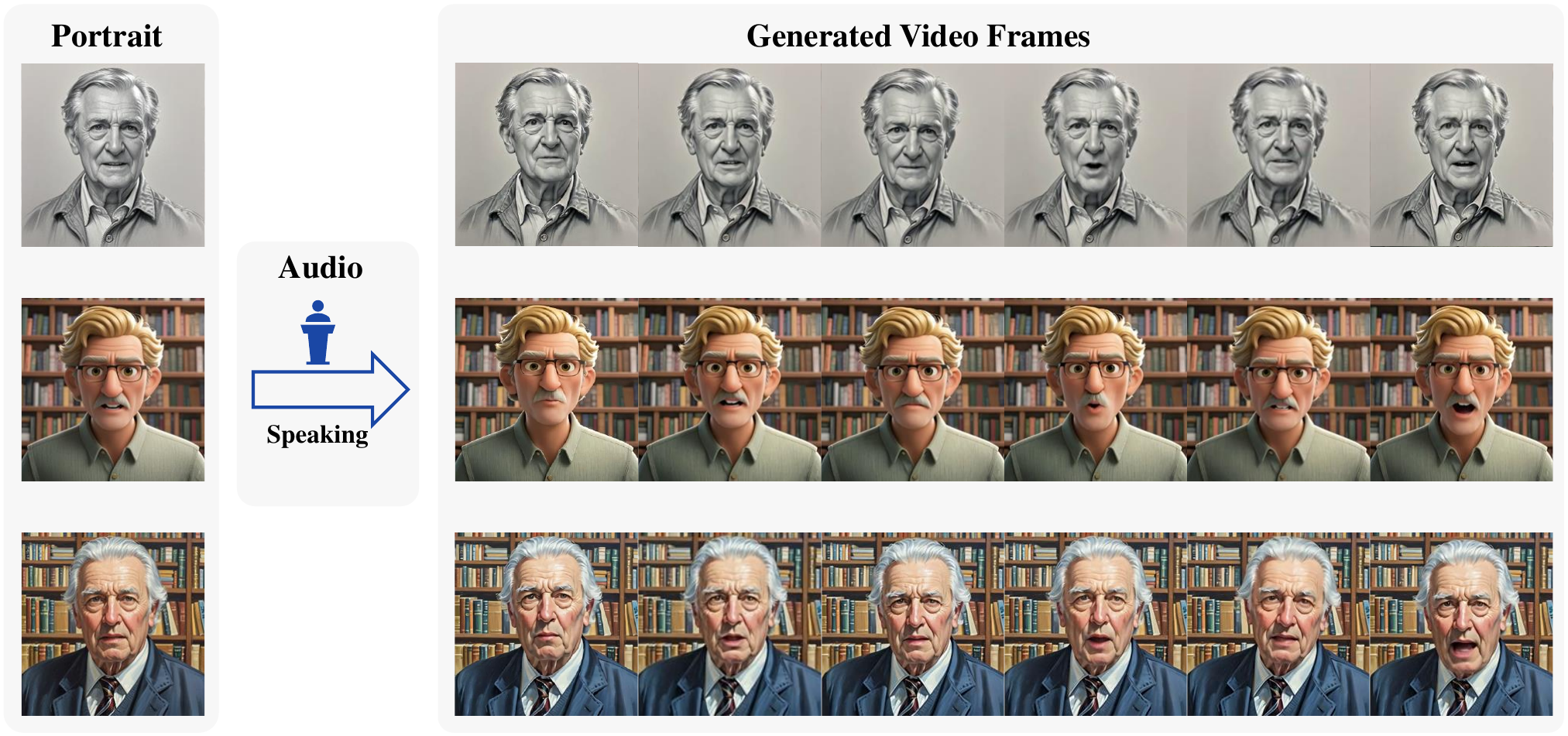}
  \caption{Video generation results of EmotiveTalk given different portraits and same speech audio signal.}
  \label{fig:multispeaker}
\end{figure*}

\noindent\textbf{Results on Multiple Portrait Styles.} To further validate the generalization capability of our proposed method across different portrait styles, we use the same speech audio to drive three distinct styles of portraits: photorealistic, cartoon, and sketch, and generate speech videos corresponding to each portrait. To compare the results, we visualize frames captured at the same time stamps from the generated videos, as shown in~\cref{fig:multispeaker}. The results demonstrate that our method successfully produces realistic speaking videos for all three styles. Moreover, the lip movements across the three videos remain highly consistent at corresponding time points, confirming the excellent generalization performance of our method across diverse portrait styles.

\begin{figure*}[t]
  \centering
  \includegraphics[width=1.0\linewidth]{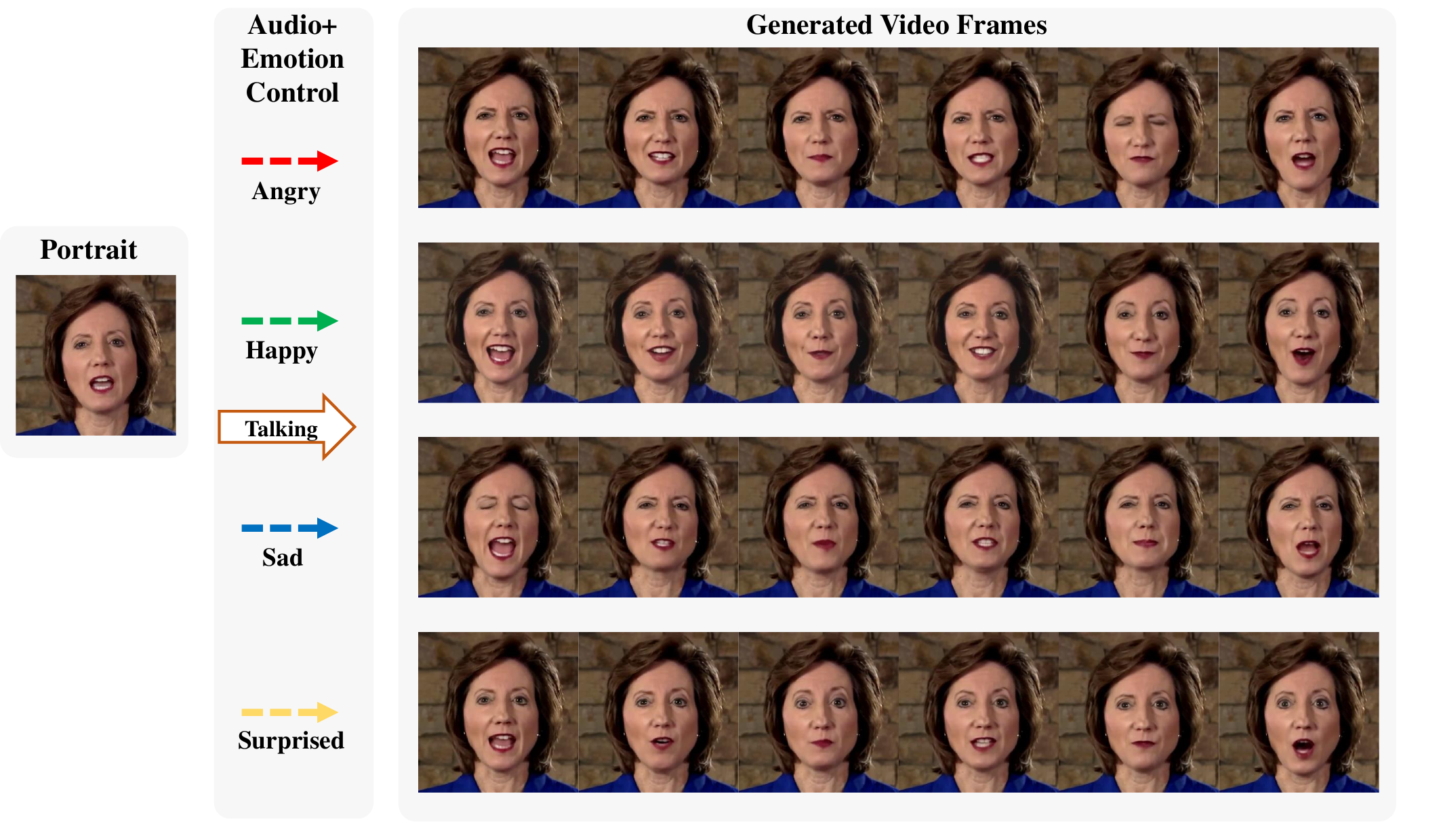}
  \caption{Video generation results of EmotiveTalk given same portraits and same speech audio signal but different emotion control conditions to generate different emotions.}
  \label{fig:multiemotion}
\end{figure*}

\noindent\textbf{Multiple Languages.} Supplementary video shows that our method generates satisfactory results with speech in French, Chinese, English, and even minority language like Cantonese. This is primarily attributed to the pre-trained wav2vec encoder's strong generalization capability across different languages, which enhances the versatility of the EmotiveTalk framework in generating talking head videos across diverse linguistic contexts

\noindent\textbf{Results on Multiple Emotion Generation.} To further evaluate the expressive capability of our proposed method in controlling and generating different emotional states, we applied the emotion control pipeline introduced in~\cref{sec:MEC}. Using the same portrait and audio input, we controlled the output by specifying different emotional states via text-based control condition. Four common emotional states, including angry, happy, sad, and surprised, are generated in our experiment. Frames captured at the same time stamps from these videos are visualized for comparison, as shown in~\cref{fig:multiemotion}.

The results demonstrate that our method can produce vivid, natural, and realistic emotional expressions. Additionally, the lip movements across the videos remain highly consistent at corresponding time points under varying emotional conditions. This confirms that our method effectively decouples the influence of facial expressions and lip synchronization, allowing for accurate lip movements synchronized with speech while transferring emotional states.

\section{Limitations and Future Work}
Despite EmotiveTalk's promising advancements in expressive talking head generation and emotion control ability, it still encounters several challenges that open the way for future research.

First, EmotiveTalk occasionally experiences motion blur during significant body movements or dramatic facial expression changes, which reduces the resolution of the generated video. This issue is associated with motion blur in the training data during periods of intense movement. A potential solution could involve applying motion blur detection and video processing techniques to the original training videos to eliminate motion blur.

Second, as there are currently no publicly available talking head video datasets with textual annotations, EmotiveTalk currently supports a limited range of emotion categories controlled via textual input, focusing on discrete emotional states. Future work could explore temporally annotating fine-grained facial expressions and emotion states in the training data using Multimodal Large Language Models (MLLMs) to build a 
talking head dataset with fine-grained textual annotations to support the research on text-guided finer-grained emotional control.

Lastly, EmotiveTalk presently focuses exclusively on controlling emotion states in talking head video generation and does not incorporate explicit control over head movements. In the generated videos, head motion primarily arises from the sampling of the diffusion model. Future enhancements could include explicit control signals for head movement, enabling precise manipulation of desired head motion patterns.

Despite these challenges, EmotiveTalk demonstrates exceptional performance and application potential in generating stable videos with expressive and controllable facial expressions. It holds significant academic and practical value, providing a foundation for future research in the field of talking head generation.

\section{Ethical Consideration}

EmotiveTalk is capable of generating highly realistic talking head videos that are difficult to distinguish from genuine footage, endowing it with extensive practical value. While EmotiveTalk holds significant positive implications for social development and technological advancement, assisting professionals in practical domains such as human-computer interaction, remote education, and caregiving companionship, the potential misuse of EmotiveTalk could lead to the spread of misinformation. For instance, it could be exploited to create fake videos using the portraits of celebrities, produce videos containing sexual innuendo or violent content, or generate counterfeit videos for purposes of extortion. Such misuses may result in substantial negative impacts, which contravene our fundamental intent of leveraging artificial intelligence to enhance human creativity, drive technological progress, and improve our society. These are outcomes we find absolutely intolerable.

We have taken the risk of potential misuse into careful consideration throughout the development of EmotiveTalk. During the training phase, we meticulously curated the training data to rigorously exclude any undesirable content involving violence, sexual implications, or horror themes. Furthermore, we have imposed strict limitations on the use of EmotiveTalk. The version deployed for academic research purposes is under the supervision of our risk assessment team. All images and audio inputs used to generate talking head videos undergo stringent evaluation and review to ensure that EmotiveTalk is not utilized to produce inappropriate information or content. For potential future releases of EmotiveTalk models intended for engineering applications, we also plan to implement a stringent review and assessment process to guarantee that generated content remains free of harmful materials. Moreover, we advocate for research on advanced forgery detection techniques, which can identify synthetic fake images and videos, thereby helping to mitigate illegal use. We remain resolute in our commitment to preventing the generation of harmful content and mitigating adverse societal impacts, and we will fully address and prevent the various potential misuses of EmotiveTalk.

% \section{Rationale}
% \label{sec:rationale}
% % 
% Having the supplementary compiled together with the main paper means that:
% % 
% \begin{itemize}
% \item The supplementary can back-reference sections of the main paper, for example, we can refer to \cref{sec:intro};
% \item The main paper can forward reference sub-sections within the supplementary explicitly (e.g. referring to a particular experiment); 
% \item When submitted to arXiv, the supplementary will already included at the end of the paper.
% \end{itemize}
% % 
% To split the supplementary pages from the main paper, you can use \href{https://support.apple.com/en-ca/guide/preview/prvw11793/mac#:~:text=Delete%20a%20page%20from%20a,or%20choose%20Edit%20%3E%20Delete).}{Preview (on macOS)}, \href{https://www.adobe.com/acrobat/how-to/delete-pages-from-pdf.html#:~:text=Choose%20%E2%80%9CTools%E2%80%9D%20%3E%20%E2%80%9COrganize,or%20pages%20from%20the%20file.}{Adobe Acrobat} (on all OSs), as well as \href{https://superuser.com/questions/517986/is-it-possible-to-delete-some-pages-of-a-pdf-document}{command line tools}.